\documentclass[11pt,twoside,twocolumn]{article}

\usepackage[T1]{fontenc} 
\usepackage[utf8]{inputenc}  
\usepackage[english]{babel}

\usepackage[hmarginratio=1:1,outer=19.4mm,top=20mm,bottom=28mm]{geometry} 
\usepackage{natbib}
\usepackage{microtype} 
\usepackage{times}
\usepackage{helvet}
\usepackage{courier}
\usepackage{url}
\usepackage[usenames,dvipsnames]{xcolor}
\usepackage[ocgcolorlinks,pdfpagemode={UseOutlines},bookmarks=true,bookmarksopen=true, bookmarksopenlevel=0,bookmarksnumbered=true,hypertexnames=false, hidelinks, pdfstartview={FitV},unicode,breaklinks=true,colorlinks=true, citecolor=darkblue, linkcolor=BrickRed, urlcolor=Blue]{hyperref}
\usepackage[raggedright]{titlesec}
\usepackage{flushend}
\usepackage{appendix}
\usepackage{chngcntr} 

\usepackage{wrapfig}
\usepackage{capt-of}
\usepackage{subcaption}
\usepackage{sidecap}
\usepackage{caption}
\usepackage{graphicx}
\usepackage{booktabs}
\usepackage{multirow}
\usepackage{slashbox}
\usepackage{makecell}
\usepackage{adjustbox}
\usepackage[ruled,vlined]{algorithm2e}
\usepackage{enumitem}
\usepackage{amsmath}
\usepackage{newtxtext} 
\usepackage{newtxmath} 
\usepackage{mathtools}
\usepackage{siunitx}
\usepackage[eulergreek]{sansmath} 
\usepackage{tikz}
\usetikzlibrary{decorations.pathreplacing}
\usetikzlibrary{decorations.markings}
\usetikzlibrary{patterns}
\usetikzlibrary{positioning}
\usetikzlibrary{shapes,arrows}
\usetikzlibrary{calc}
\usetikzlibrary{intersections}
\usetikzlibrary{plotmarks}
\usetikzlibrary{spy,backgrounds}
\usepackage{pgfplots}
\usepgflibrary{plotmarks}
\usepgfplotslibrary{units}
\pgfplotsset{compat=1.12}
\usetikzlibrary{pgfplots.groupplots}
\usetikzlibrary{external}
\usepackage{comment}

\newcommand{\algBreak}{\\[1mm]}
\newcommand{\algro}[2]{\ \\[-4.7mm] \hfill \parbox[t]{1.0cm+#1}{#2}}

\DeclareMathOperator{\argmin}{argmin} 

\newcommand{\disT}{\textstyle}

\usepackage{color}

\definecolor{white}{rgb}{1.0,1.0,1.0}
\definecolor{brightred}{rgb}{1.0,0.1,0.1}
\definecolor{brightblue}{rgb}{0.0,0.0,0.8}
\definecolor{darkblue}{rgb}{0.0,0.0,0.5}
\definecolor{darkgreen}{rgb}{0.0,0.3,0.0}
\definecolor{brightgreen}{rgb}{0.0,0.8,0.0}
\definecolor{darkblack}{rgb}{0.0,0.0,0.0}
\definecolor{grey}{rgb}{0.3,0.3,0.3}

\newcommand{\mycomment}[1]{}

\title{\vspace{-32pt}\LARGE Large Scale Clustering with Variational EM \\ for Gaussian Mixture Models}

\author{
\large
Florian Hirschberger, Dennis Forster, J\"org L\"ucke \\
\normalsize florian.hirschberger@uol.de, dennis.forster@uol.de, joerg.luecke@uol.de\\
\normalsize Machine Learning Lab, University of Oldenburg, Germany
}
\date{}

\begin{document}

\twocolumn[
\begin{@twocolumnfalse}
\maketitle
\begin{abstract}
\noindent This paper represents a preliminary (pre-reviewing) version of a sublinear variational algorithm for isotropic Gaussian mixture models (GMMs).
Further developments of the algorithm for GMMs with diagonal covariance matrices (instead of isotropic clusters) and their corresponding benchmarking results have been published by TPAMI (\href{https://doi.org/10.1109/tpami.2021.3133763}{doi:10.1109/TPAMI.2021.3133763}) in the paper ``A Variational EM Acceleration for Efficient Clustering at Very Large Scales''.
We kindly refer the reader to the TPAMI paper instead of this much earlier arXiv version (the TPAMI paper is also open access).
Publicly available source code accompanies the paper (see \url{https://github.com/variational-sublinear-clustering}).
Please note that the TPAMI paper does not contain the benchmark on the 80~Million Tiny Images dataset anymore because we followed the \href{https://groups.csail.mit.edu/vision/TinyImages/}{call of the dataset creators} to discontinue the use of that dataset.

The aim of the project (which resulted in this arXiv version and the later TPAMI paper) is the exploration of the current efficiency and large-scale limits in fitting a parametric model for clustering to data distributions.
To reduce computational complexity, we used a clustering objective based on truncated variational EM (which reduces complexity for many clusters) in combination with coreset objectives (which reduce complexity for many data points).
We used efficient coreset construction and efficient seeding to translate the theoretical sublinear complexity gains into an efficient algorithm.
In applications to standard large-scale benchmarks for clustering, we then observed substantial wall-clock speedups compared to already highly efficient clustering approaches.
To demonstrate that the observed efficiency enables applications previously considered unfeasible, we clustered the entire and unscaled 80~Million Tiny Images dataset into up to 32,000 clusters.
\end{abstract}
\vspace{20pt}
\end{@twocolumnfalse}
]

\section{Introduction}
\label{sec:Introduction}

Efficiency demands for clustering algorithms are constantly increasing.
As the volume of data and the number of data-driven tasks constantly grows, new algorithms for increasingly many clusters \citep{PellegMoore1999,CoatesEtAl2011,Curtin2017,KobrenEtAl2017,NechEtAl2017,OttoEtAl2018} on increasingly large data sets \citep{HilbertLopez2011} are required.
The execution of standard clustering algorithms, such as $k$-means or expectation maximization (EM) for Gaussian mixture models (GMMs) \citep{McLachlanPeel2004}, quickly becomes prohibitive at large scales as their run-time costs grow with $\mathcal{O}(NCD)$ per iteration (where $N$ is the number of data points, $C$ the number of clusters, and $D$ the dimensionality of the data points).
For any combination of large $N$, $C$, and $D$, the execution of even just one iteration may exceed the limits of state-of-the-art computational hardware.

\paragraph{Related Work and Own Contribution}
%
Separate lines of research address the demand for more efficient clustering algorithms in different ways.
Typically the focus is either on reducing the dependency on $N$, on $C$, or on $D$ individually, or to reduce the required number of learning iterations:
(1)~Research on coresets \citep{HarPeledMazumdar2004} aims to reduce the linear dependency on $N$ by replacing the full data set by a weighted subset. 
(2)~Novel variational EM approaches aim to reduce the linear dependency on $C$ by replacing full E-steps by more efficient partial E-steps~\citep{ForsterLucke2018}.
Alternatively, construction and pruning of dual K-D-tree data structures~\citep{Curtin2017} can also reduce the linear dependency on~$C$.
(3)~The dependency on $D$ has been addressed relatively early using properties of the triangle inequality \citep{Elkan2003} or random projections \citep{ChanLeung2017}.
Finally, (4)~the number of iterations required until convergence can be reduced by advanced parameter initializations (seeding) \citep{ArthurVassilvitskii2007,BachemEtAl2016b,BachemEtAl2016a,NewlingFleuret2017}.
\break\indent
In this work we combine different lines of research. A key role is played by variational EM (sometimes referred to as Variational Bayes), which replaces the log-likelihood learning objective
by a lower bound that can be optimized more efficiently.
Such variational lower bounds go under the name of {\em variational free energies} \citep{NealHinton1998,JordanEtAl1999} or {\em evidence lower bounds} \citep[ELBO; e.g.,][]{HoffmanEtAl2013}
and different choices of variational distributions result in different lower bounds.
Here we use truncated variational approximations, which have been applied to different types of data models, including sparse coding
\citep[e.g.,][]{SheikhLucke2016}, topic models \citep{HughesSudderth2016}, and mixture models \citep{DaiLucke2014,SheltonEtAl2017,HughesSudderth2016,LeeEtAl2017}.
Truncated approximations increase efficiency by neglecting latent values with low posterior probabilities.
For clustering with isotropic clusters, this means neglecting distant clusters \citep{LuckeForster2019}.
Neglection ideas have, in general, been observed to reduce computational demands for probabilistic clustering approaches \citep{DaiLucke2014,HughesSudderth2016,ForsterLucke2017} as well as for deterministic approaches such as $k$-means or agglomerative clustering \citep[e.g.][]{NechEtAl2017,OttoEtAl2018}.
For $k$-means, e.g.\ \citet{Phillips2002,AgustssonEtAl2017} obtained algorithms scaling with $\mathcal{O}(N\gamma{}D+C^2D)$ per iteration (where $\gamma<C$).
In contrast, other popular approaches for the acceleration of clustering \citep{Elkan2003,Hamerly2010} require at least one iteration which scales with $\mathcal{O}(NCD)$ (plus an $\mathcal{O}(C^2D)$ term to, e.g., keep track of boundary values for distances).
Important for this work, efficiency of variational EM iterations can be substantially further increased by applying partial instead of full variational E-steps to GMMs \citep{ForsterLucke2018}.
The most efficient such algorithms required $\mathcal{O}(NG^2D)$ computations per iteration where $G\ll{}C$ is a small constant depending on cluster neighborhood relations.
Variational EM also means an increase in required memory, however.
While substantially fewer distance evaluations where observed until convergence \citep{ForsterLucke2018}, it remained an open question how significant the computational overhead, e.g.\ for memory access, impacts practical run time performance.
\break\indent
In this work we combine different lines of research.
Our main goal is (1st)~to explore the efficiency and large scale limits in fitting mixture models to data.
Our specific focus are data sets with large $N$, $C$, and $D$ as they arise increasingly frequently \citep{TorralbaEtAl2008,NechEtAl2017,OttoEtAl2018}.
As standard approaches are not sufficiently efficient, data analysis algorithms therefore often use task specific mechanisms to explore cluster-to-data neighborhoods including learning smaller sets of features before clustering, the design efficient task-specific metrics, or greedy approaches for agglomerative clustering \citep{NechEtAl2017,OttoEtAl2018}.
In contrast, here our focus is to approach the problem head-on by directly fitting a GMM (with isotropic clusters) to data, i.e., we follow the standard goal of fitting a probabilistic model to a data distribution.
The substantial efficiency increases and scaling behavior we observe for our approach, and the novel large scale limits we report, may be considered as the main contribution of the paper.
\break\indent
To reach our main goal, we first merge fast coreset approximations \citep{BachemEtAl2018} with fast variational EM approximations \citep{ForsterLucke2018} for GMMs.
Here our contributions are
(2nd)~the derivation of a mathematically grounded approach to derive a single clustering objective which combines variational lower bounds with coreset likelihoods, and
(3rd)~a concrete realization of an algorithm able to translate the theoretical efficiency gains into substantially reduced run times for clustering.
A consequence of the high efficiency in optimizing the merged coreset/variational EM objective is that standard coreset construction and conventional seeding \citep[e.g.\ of $k$-means++,][]{ArthurVassilvitskii2007} become bottlenecks.
We therefore apply the most efficient current algorithm for coreset construction \citep{BachemEtAl2018} and seeding \citep{BachemEtAl2016b}.
Both methods are reimplemented, optimized and synchronized with the core iterative parameter updates, which may be considered our (4th) contribution.

\section{Merged objective and efficient optimization}
\label{sec:MergedObjective}

We approach the clustering task by fitting a probabilistic model in form of Gaussian mixtures to a set of $N$~data points $\vec{y}^{\,(1)},\dots,\vec{y}^{\,(N)}\in\mathbb{R}^D$.
We use the most elementary such mixture model with $C$~isotropic clusters and equal mixing proportions.
For a data point $\vec{y}$, the model's probability density is then given by
\begin{align}
&p(\vec{y}|\Theta) = \frac{1}{C} (2\pi\sigma^2)^{-\frac{D}{2}} \sum^C_{c=1}\! \exp\!\Big(\!-\frac{1}{2\sigma^2}\|{}\vec{y}-\vec{\mu}_c\|{}^2\Big),
\label{eq:GMMIso}
\end{align}
where $\|{}\cdot\|$ denotes the Euclidean distance in $\mathbb{R}^D$.
Given this model, clustering takes the form of finding $C$~cluster centers~$\vec{\mu}_c$ and one cluster variance~$\sigma^2$, which we denote by $\Theta=(\vec{\mu}_1,\ldots,\vec{\mu}_C,\sigma^2)$.

\paragraph{Merging Coreset and Variational Objective}
%
On a set of $N$ data points, a coreset can be defined as a subset of $N'<N$ data points with $N'$ positive weights $\gamma^{(n)}\in\mathbb{R}^{+}$ such that training on the coreset approximates training on the original data set.
Given such a coreset $\{\vec{y}^{\,(n)},\gamma^{(n)}\}_{n=1}^{N'}$ and GMM (\ref{eq:GMMIso}), we can, following \citet{LucicEtAl2018}, define a log-likelihood on the full data and on the coreset of the following form:
\vspace{-4pt}
\begin{equation}
\begin{aligned}
L(\Theta) &= \sum_{n=1}^{N} \log\!\Big( p(\vec{y}^{\,(n)}|\Theta)\Big),\\
L^{\mathrm{core}}(\Theta) &= \sum_{n=1}^{N'} \gamma^{(n)}\log\!\Big( p(\vec{y}^{\,(n)}|\Theta)\Big),
\label{eq:LikelihoodCore}
\end{aligned}
\end{equation}
where $p(\vec{y}|\Theta)$ denotes the GMM density~(\ref{eq:GMMIso}).
Standard coreset constructions ensures that the optimization result for $L^{\mathrm{core}}(\Theta)$
closely approximates the optimization result for $L(\Theta)$ \citep[e.g.,][]{LucicEtAl2018}.

A different approximation of $L(\Theta)$ is given by variational EM \citep[e.g.,][]{JordanEtAl1999}, which optimizes a lower bound of the log-likelihood.
Such variational lower bounds are defined to be efficiently computable while matching the actual log-likelihood as closely as possible.
Variational bounds, here denoted by $\mathcal{F}(\Lambda,\Theta)$, contain additional parameters $\Lambda$, which are optimized in variational E-steps that replace the computation of full posterior probabilities in standard E-steps.

We now combine approximation $L^{\mathrm{core}}(\Theta)$ with efficiently computable variational bounds $\mathcal{F}(\Lambda,\Theta)$.
These bounds result from the introduction of variational approximations $q^{(n)}(c;\Lambda)$ to exact
posteriors $p(c|\vec{y}^{\,(n)},\Theta)$ via the application of Jensen's inequality \citep{Jensen1906,Bishop2006}.
For our purposes, we choose variational distributions in the form of truncated posteriors with variational parameters $\Lambda=(\mathcal{K},\hat{\Theta})$:
\vspace{-3pt}
\begin{align}
s_{c}^{(n)} \! \vcentcolon= q^{(n)}(c;\mathcal{K},\hat{\Theta}) \! = \! \frac{p(c,\vec{y}^{\,(n)}|\hat{\Theta})}{\sum_{\tilde{c} \in \mathcal{K}^{(n)}} \! p(\tilde{c},\vec{y}^{\,(n)}|\hat{\Theta})} \delta(c\!\in\!\mathcal{K}^{(n)}),
\label{eq:TruncMain}
\end{align}
where $\hat{\Theta}$ are the cluster means and variance of the variational distribution, $\mathcal{K} = \{\mathcal{K}^{(1)},\dots,\mathcal{K}^{(N)}\}$ are $N$~subsets of the index set $\{1,\ldots,C\}$, and $\delta(c\in\mathcal{K}^{(n)})=1$ if $c\in\mathcal{K}^{(n)}$ and zero otherwise.
Truncated posteriors are a natural choice to consider more than one `winning' cluster while otherwise maintaining `hard' zeros \citep{ForsterLucke2018,LuckeForster2019}.
The size of all $\mathcal{K}^{(n)}$ we take to be constrained to $|\mathcal{K}^{(n)}|=C'$ (i.e., $C'$ `winning' clusters).
If full posteriors are dominated by few large values, distributions~(\ref{eq:TruncMain}) approximate full posteriors very well.
For clustering of natural data, such dominance of few posterior values is indeed very common.
While closely approximating true posteriors, distributions~(\ref{eq:TruncMain}) were also shown to significantly reduce computational costs in~$C$ \citep{DaiLucke2014,SheikhLucke2016,HughesSudderth2016}.
For coreset data, reducing $N$ to $N'$, a variational bound with truncated distributions can then be derived to have the following form (details in Suppl.\,\ref{app:FreeEnergy}):
\vspace{-3pt}
\begin{equation}
%
\mathcal{F}(\mathcal{K},\hat{\Theta},\Theta)
 = \! \sum\limits_{n=1}^{N'} \!\gamma^{(n)} \!\!\!\! \sum_{c \in \mathcal{K}^{(n)}} \!\!\!\! s_{c}^{(n)} \log\!\Big( \frac{p(c,\vec{y}^{\,(n)}|\Theta)}{s_{c}^{(n)}}\Big).
\label{eq:FreeEnergyExplicit}
\end{equation}
The total number of terms that have to be computed for objective (\ref{eq:FreeEnergyExplicit}) now grows with $\mathcal{O}(N'{}C')$, where \mbox{$C'\leq C$} is the number of clusters considered for each data point.
The relation between the original, coreset, and variational objective can be summarized as follows:
\vspace{-1pt}
\begin{align}
L(\Theta) \approx L^{\mathrm{core}}(\Theta) \geq \mathcal{F}(\mathcal{K},\hat{\Theta},\Theta).
\label{eq:ApproxRelations}
\end{align}
The cost to compute the objectives decreases from left to right, but also the approximation quality decreases.
However, while the computational cost strongly decreases, we can expect to maintain a relatively high approximation quality.
A number of results have shown that coresets give rise to very accurate clustering results compared to full data sets \citep{LucicEtAl2018,BachemEtAl2018}, and truncated variational EM has been shown to result in tight lower bounds and to be advantageous in avoiding local optima \citep{SheikhEtAl2014,SheikhLucke2016,HughesSudderth2016,ForsterLucke2018}.
Furthermore, the degree of accuracy and efficiency can be traded off by choosing the approximation parameters $C'$ and $N'$.
In their limits, the variational lower bound recovers the coreset likelihood (for $C'\rightarrow{}C$), and coreset likelihoods typically recover the original likelihood for $N'\rightarrow{}N$.
The question now remains, how objective~(\ref{eq:FreeEnergyExplicit}) can be optimized as efficiently as possible.

\paragraph{Optimization of Model Parameters: M-Step}
%
Although the number of summands in the merged objective~(\ref{eq:FreeEnergyExplicit}) is strongly reduced, its basic analytical structure remains similar to the standard structure of variational lower bounds.
The derivation of parameter updates for (\ref{eq:FreeEnergyExplicit}) can therefore proceed essentially along the same lines as standard EM for GMMs.
By following \citet{LucicEtAl2018} and by simultaneously replacing full posteriors $p(c|\vec{y}^{\,(n)},\Theta)$ with the variational distributions $s_{c}^{(n)}$ of Eq.\,(\ref{eq:TruncMain}), we obtain:
\begin{equation}
\begin{aligned}
\vec{\mu}^{\mathrm{\,new}}_c &=  \frac{\sum_{n=1}^{N'}\gamma^{(n)}s_{c}^{(n)}\vec{y}^{\,(n)}}{\sum_{n=1}^{N'}\gamma^{(n)}s_{c}^{(n)}}, \\
\sigma_{\mathrm{new}}^2 &= \frac{1}{D \sum_{n=1}^{N'}\gamma^{(n)}}\sum_{n=1}^{\;N'}\sum_{\,c\in\mathcal{K}^{(n)}\!\!\!}\gamma^{(n)} s_{c}^{(n)} \|\vec{y}-\vec{\mu}^{\,\mathrm{new}}_c\|^2. \hspace{-20pt}
\label{eq:GMMMStep}
\end{aligned}
\end{equation}
Because of the `hard' zeros of $s_{c}^{(n)}$ for $c\not\in\mathcal{K}^{(n)}$, the number of M-step computations is of $\mathcal{O}(N'C'D)$.

\paragraph{Optimization of Variational Parameters: E-Step}
%
A crucial and often computationally very demanding step in optimizations of probabilistic generative models is the E-step.
In our case, we seek variational parameters~$(\mathcal{K},\hat{\Theta})$ that optimize the variational bound~(\ref{eq:FreeEnergyExplicit}) while keeping $\Theta$ fixed.
The size of the search space to find the best $\mathcal{K}^{(n)}$ is $\binom{\!C}{\,C'}{}\!$ for each $n$. A concern
may therefore be that any complexity reduction in the evaluation of objective (\ref{eq:FreeEnergyExplicit}) or in the corresponding learning equations are dominated by the combinatorial problem in finding $\mathcal{K}= \{\mathcal{K}^{(1)}, \dots, \mathcal{K}^{(N')}\}$.
In general, optimizations of functions depending on $\mathcal{K}$ can indeed not be expected to be very efficient.
For the merged objective~(\ref{eq:FreeEnergyExplicit}), we can however make use of a number of theoretical results for truncated variational distributions, as discussed in detail in Suppl.\,\ref{app:VarEStep}.
A main observation is, that it turns out to be sufficient to find for each $\mathcal{K}^{(n)}$ the $C'$ clusters $c$ with the largest joints $p(c,\vec{y}^{\,(n)}|\Theta)$ to maximize (\ref{eq:FreeEnergyExplicit}).
The optimization problem is consequently solvable with $\mathcal{O}(N'C)$ computations in total.
Such a scaling is much more favorable than the combinatorics suggested.
However, compared to the M-step with $\mathcal{O}(N'C')$, an $\mathcal{O}(N'C)$ scaling is still not efficient enough.
Therefore, we further improve efficiency by merely seeking to {\em increase} objective~(\ref{eq:FreeEnergyExplicit}) in each E-step instead of fully maximizing it.
To do so, we generalize the result of \citet{ForsterLucke2018} for variational bounds with truncated distributions (\ref{eq:TruncMain}) to coreset-weighted truncated variational bounds, i.e., we show that pair-wise comparisons of distances are sufficient to warrant the variational bound to increase:

\paragraph{Proposition 1}
%
Consider a coreset $(\vec{y}^{\,(n)},\gamma^{(n)})_{n=1\dots N'}$ for the GMM~(\ref{eq:GMMIso}) with parameters $\Theta$, and the merged variational bound~(\ref{eq:FreeEnergyExplicit}) with variational parameters $\mathcal{K}$.
If we replace for an arbitrary $n$ a cluster $c\in\mathcal{K}^{(n)}$ by a cluster $c^{\mathrm{new}}\not\in\mathcal{K}^{(n)}$, then the variational bound increases if and only if:
\begin{align}
\|\vec{y}^{\,(n)}-\vec{\mu}_{c^{\mathrm{new}}}\| < \|\vec{y}^{\,(n)}-\vec{\mu}_{c}\|
\label{eq:Criterion}
\end{align}
The proof is given in Suppl.\,\ref{app:ProofPropOne}. \hfill $\square$

The important result is that the coreset weights $\gamma^{(n)}$ do not change criterion~(\ref{eq:Criterion}) compared to results for full data sets.
This is because sets $\mathcal{K}^{(n)}$ are optimized individually for each~$n$, making the pair-wise comparison of distances unaffected by the common positive multiplier~$\gamma^{(n)}$.
We can consequently apply the same efficient variational E-step as suggested for the \mbox{{\em var-GMM-S}} algorithm  by \citet{ForsterLucke2018}.

\section{Realization of the complete algorithm}
\label{sec:AlgRealization}

A concrete algorithm applicable to very large scales has to combine different modules of which the derived optimization of objective (\ref{eq:FreeEnergyExplicit})
is an essential component but just one of the following:

\paragraph{Constructing Lightweight Coresets}
%
The first module of the complete algorithm is the construction of a coreset of $N'$ data points.
Coresets come with different theoretical guarantees and scaling properties.
Previously suggested coreset constructions \citep{HarPeledMazumdar2004,LucicEtAl2018} allow for relatively small coresets with for usual scales relatively small construction costs.
For the scales we are interested in, standard coreset construction would become a computational bottleneck.
We therefore choose the recently suggested lightweight coresets \citep{BachemEtAl2018} (LWCS), which are substantially faster to construct.
They require only two passes through the data set with complexity $\mathcal{O}(ND)$.

\paragraph{Efficient Seeding}
%
The second module is an efficient seeding approach.
Novel seeding algorithms \citep{BachemEtAl2016b,BachemEtAl2016a,NewlingFleuret2017} can significantly improve the clustering quality and reduce the number of iterations until convergence. 
Since standard seeding methods are no longer feasible for large $N$ and large $C$ (e.g., $k$-means++ seeding scales with $\mathcal{O}(NCD)$), we adopt highly efficient AFK-MC$^2$ seeding \citep{BachemEtAl2016b}, which has also been used for \mbox{{\em var-GMM-S}} in \citet{ForsterLucke2018}.
For this seeding, an initial single pass through the coreset data with $\mathcal{O}(N'D)$ is required to define a proposal distribution. 
The cluster centers are then computed using independent Markov chains of length $m$, resulting in a main seeding loop with complexity $\mathcal{O}(mC^{2}D)$.

\paragraph{Initial variance estimation}
%
In addition to the provided initial cluster centers $\vec{\mu}_{1:C}$, we require in the third module an efficient estimation of the initial variance $\sigma^2$.
As an exact variance calculation based on initial centers and data points would scale with $\mathcal{O}(NCD)$, we have to find an more efficient estimation.
To do so, we use that the optimization of the variational parameters $\mathcal{K}^{(n)}$ during the E-step is independent of the variance.
We can therefore first optimize $\mathcal{K}^{(n)}$ without having to compute $s_c^{(n)}$ or performing M-steps.
As the $\mathcal{K}^{(n)}$ estimates the closest cluster centers for each data point, the distances of data points $ \vec{y}^{\,(n)}$ with regard to clusters in $\mathcal{K}^{(n)}$ then provide an estimate of the data variance of $\mathcal{O}(N'C'D)$:
\vspace{-8pt}
\begin{align}
\disT\sigma_\mathrm{init}^{2} = \sum_{n=1}^{N'} \frac{\gamma^{(n)} \min_{c \in \mathcal{K}^{(n)}} \|\vec{\mu}_{c,\mathrm{init}} - \vec{y}^{\,(n)}\|^2}{D \sum_{n=1}^{N'}\gamma^{(n)}}.
\label{eqn:sigma}
\end{align}
In our experiments, the variational algorithms are observed to be very robust to these initial values, with a coarse estimate of the approximate order of magnitude generally being sufficient.
Thus, no additional $\mathcal{K}^{(n)}$ optimization steps other than the first E-step (or initial E-steps as discussed in Suppl.~\ref{app:Exp}) were necessary for our experiments to gain
sufficient initial values $\sigma^2$.

\begin{algorithm}[h]
construct LWCS $(\vec{y}^{\,(n)},\gamma^{(n)})_{n=1\dots N'}$;\\[1mm]
init $\mathcal{G}_{c}$ randomly for all $c=1,\ldots,C$;\\ 
init $\mathcal{K}^{(n)}$ randomly for all $n=1,\ldots,N'$;\\ 
init $\vec{\mu}_{1:C}$ (with AFK-MC$^2$ seeding);\\
init $\sigma^2$ with Eq.\,\ref{eqn:sigma} after 1st $\mathcal{K}^{(n)}$-update;\\[1mm]
%
%
\Repeat{$\mathcal{F}(\vec{\mu}_{1:C},\sigma^2)$ has converged}
{
	\For{$n=1:N'$}{
			$\mathcal{G}^{(n)} =\cup_{c\in\mathcal{K}^{(n)}}\mathcal{G}_{c}$; \\
			\For{$c\in\mathcal{G}^{(n)}$}{
					$d^{(n)}_{c} = \|\vec{y}^{\,(n)}\,-\,\vec{\mu}_{c}\|$;
			}
			$\mathcal{K}^{(n)} \leftarrow$ indices of $C'$ smallest $d^{(n)}_{c}$;
	 }
	 \For{$c = 1:C$}{
		 $\mathcal{G}_{c} \leftarrow$ indices of $G$ closest clusters to $c$; \hspace{-50pt}
	 }
	 \For{$n=1:N'$}{
		 \For{$c\in\mathcal{K}^{(n)}$}{
			 $s_c^{(n)} = \frac{\exp\left(-\frac{1}{2}(d_{c}^{(n)} / \sigma)^2\right)}{\sum_{c'\in\mathcal{K}^{(n)}}\exp\left(-\frac{1}{2}{(d_{c'}^{(n)} / \sigma)}^2\right)}$;\hspace{-10pt}
		 }
	 }
	 \For{$n=1:N'$}{
		 \For{$c\in\mathcal{K}^{(n)}$}{
			 $\vec{\mu}_c^{\,\mathrm{num}} = \vec{\mu}_c^{\,\mathrm{num}} + \gamma^{(n)}s_{c}^{(n)}\vec{y}^{\,(n)}$;\\
			 $\mu_c^{\,\mathrm{den}} = \mu_c^{\,\mathrm{den}} + \gamma^{(n)}s_{c}^{(n)}$;\\
		 }
	 }
	 \For{$c=1\,\ldots,C$}{
			$\vec{\mu}_c = \vec{\mu}_c^{\,\mathrm{num}} / \mu_c^{\,\mathrm{den}} $;\\
	 }
	 \For{$n=1:N'$}{
		 \For{$c\in\mathcal{K}^{(n)}$}{
			 $\sigma^2 =  \sigma^2 + \frac{\gamma^{(n)} s_c^{(n)}}{D\sum_{n}\gamma^{(n)}}  \|\vec{y}^{\,(n)}-\vec{\mu}_{c}\|^2$;\hspace{-10pt}
		 }
	 }
}


	\caption{The \textit{vc-GMM} algorithm.}
	\label{alg:vc-GMM}
\end{algorithm}

\paragraph{Variational EM for the Merged Objective.}
%
After LWCS construction (which provides $N'$ coreset data points with weights~$\gamma^{(n)}$), seeding (which provides $\vec{\mu}_{1:C}$), and variance estimation (which provides $\sigma$), we can now update the parameters to maximize objective~(\ref{eq:FreeEnergyExplicit}).
To do so, we apply parameter updates~(\ref{eq:GMMMStep}), which require the coreset weights and the variational parameters~$\mathcal{K}^{(n)}$.
By virtue of Prop.\,1, we can apply the variational loop of \mbox{\textit{var-GMM-S}} \citep{ForsterLucke2018} for the E-step, which (A)~enables
efficient optimization of the sets $\mathcal{K}^{(n)}$ with a scaling independent of $C$ and proportional to $N'$, and (B)~guarantees a monotonic increase of
objective~(\ref{eq:FreeEnergySimplified}).

\begin{figure*}
    \includegraphics[width=\textwidth]{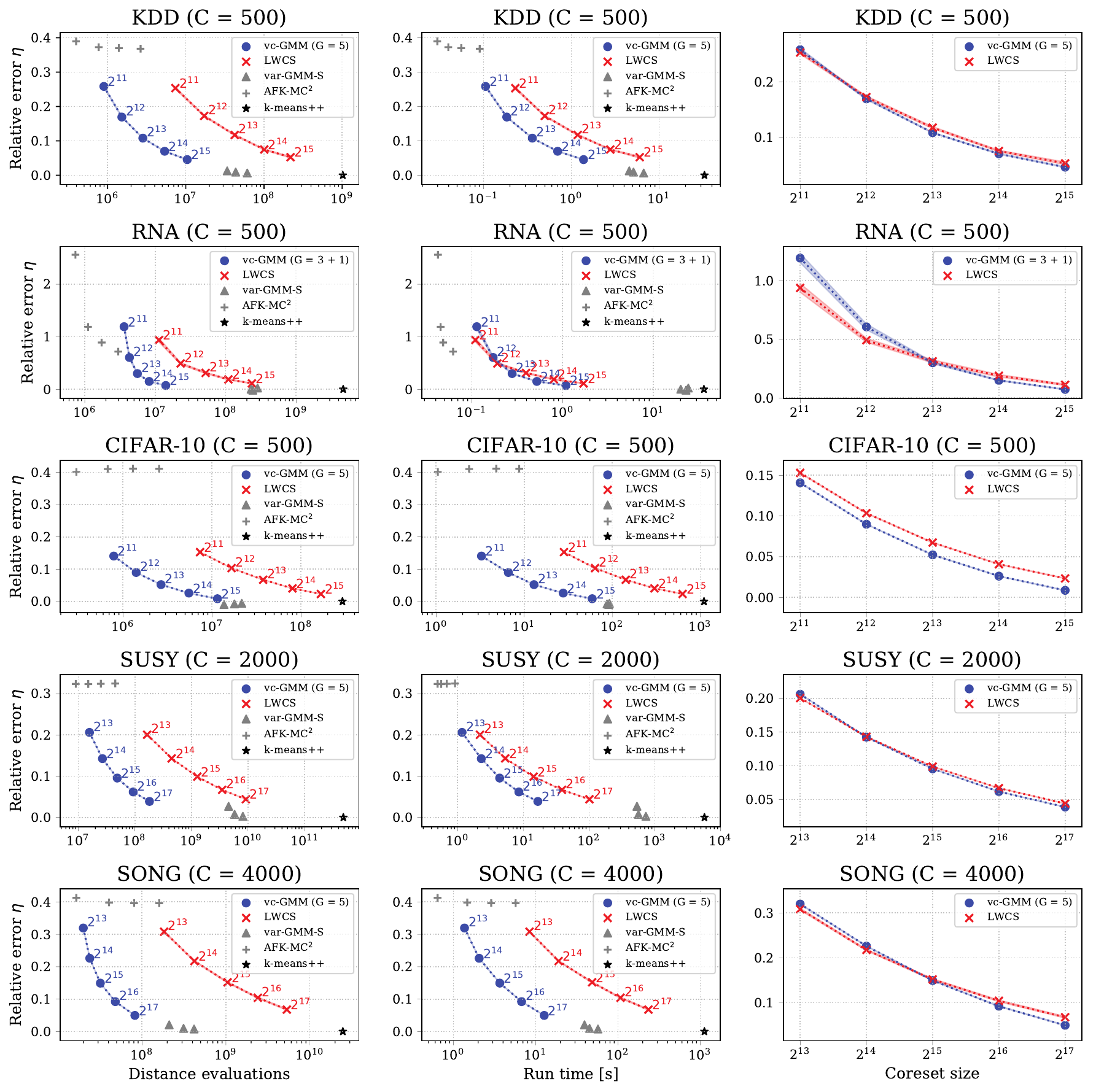}
    \caption{
			Relative quantization error and speedup of \textit{vc-GMM}, $k$-means on lightweight coresets (\textit{LWCS}), \textit{var-GMM-S}, \textit{AFK-MC$^2$} seeding, and $k$-means++.
			Each row refers to one benchmark (with increasing cluster numbers from top to bottom).
			The y-axes denote the relative quantization error with respect to $k$-means++, as given in Eq.~(\ref{eq:eta}).
			Different parameter settings of the algorithms show the trade-off between effectiveness (in terms of quantization error) and speed (in terms of distance evaluations, in the left column; and run time, in the middle column).
			Measurements for \textit{vc-GMM} (with fixed $G$ as given in the plot legends) and \textit{LWCS} are given for five different coreset sizes $N'$, denoted in the plots.
			The right column shows the performance trade-off due to decreasing coreset sizes of \textit{vc-GMM} and \textit{LWCS}.
			For \textit{vc-GMM} and \textit{LWCS} we also included the standard error of the mean (SEM), which is however very small except for one subplot (RNA).
			Measurements for \textit{var-GMM-S} refer to configurations with $G \in \{3$\,+$1, 5, 7\}$ (`+$1$' denotes one random additional cluster per $\mathcal{G}^{(n)}$), where settings with larger $G$ lie to the right, as they require more distance evaluations and higher run times.
			Measurements for \textit{AFK-MC$^2$} seeding refers to Markov chains of lengths $m=2, 5, 10, 20$ (from left to right).
			In addition to SONG with $C = 4000$, we provide SONG with $C = 2000$ in Suppl.\ Fig.\,\ref{fig:Song2k}.
		}
    \label{fig:mainFigure}
\end{figure*}

In addition to the variational parameters $\mathcal{K}$, the variational loop optimizes auxiliary index sets $\mathcal{G}_{c}$, one for each cluster $c = 1, ..., C$ and each of size $|\mathcal{G}_{c}| = G$.
Essentially, each \mbox{$\mathcal{K}^{(n)} \!\in \mathcal{K}$} comprises the $C'$ {\em estimated} closest clusters to data point $\vec{y}^{\,(n)}\!$, whereas each $\mathcal{G}_{c}$ comprises the $G$ {\em estimated} closest neighboring clusters (including itself) for each cluster~$c$.
Distances of the $N'$ data points to cluster centers $\vec{\mu}_c$ are then only evaluated over close-by neighborhood regions $\mathcal{G}^{(n)} = \bigcup_{c \in \mathcal{K}^{(n)}} \mathcal{G}_{c}$ with size of $\mathcal{O}(C'G)$ to find the $C'$ clusters with largest joints $p(c,\vec{y}^{\,(n)}|\Theta)$.
Using such reduced search regions, the used variational E-step
jointly optimizes $\mathcal{K}$ and $\{\mathcal{G}_c\}_{c=1}^{C}$, which translates in conjunction with coresets to a complexity of $\mathcal{O}(N'C'G D)$.
For simplicity, we use equal sizes for $\mathcal{K}^{(n)}$ and $\mathcal{G}_{c}$ (i.e., $C' = G$), which results in a complexity of $\mathcal{O}(N'G^{2}D)$ for each E-step.
Details about the updates of $\mathcal{G}^{(n)}$ and $\mathcal{G}_{c}$ are given in Suppl.\,\ref{app:VarLoop}.
As an optional modification, we may include one additional randomly chosen cluster to each of the search spaces $\mathcal{G}^{(n)}$ to facilitate exploration (e.g., for very small $G$), which we denote with `+1' when giving the value of $G$ (e.g.\,in Tab.\,\ref{tab:mainTable}).

The complete algorithm is summarized in Alg.\,\ref{alg:vc-GMM} and consists of all above described modules.
It will be denoted by \textit{vc-GMM} to reflect the variational and coreset components used.

\section{Numerical experiments}

\begin{table*}[tb]
    \centering
    \caption{
			Relative quantization error (with SEM) and speedup of the algorithms in Fig.\,\ref{fig:mainFigure} with $k$-means++ as baseline.
			For \textit{vc-GMM} and \textit{LWCS}, we show values for coreset sizes that result in trade-offs that are closest to a maximal increase in error relative to $k$-means++ of around $10\%$.
			For \textit{var-GMM-S}, we show values for the same $G$ parameter as used in \textit{vc-GMM}.
			The last column shows the fraction of time spent on coreset construction and seeding.
		}
		\vspace{3pt}
		\label{tab:mainTable}
		{
\fontsize{6}{6.2} \sffamily \sansmath
\renewcommand{\arraystretch}{1.1}
\setlength{\tabcolsep}{4pt}
\newcolumntype{?}[1]{!{\vrule width #1}}
\begin{tabular}{c?{0.25pt}l?{0.5pt}lll?{0.25pt}r?{0.25pt}rr?{0.25pt}rr}
		\toprule
		\textbf{Data Set} 	&           		&	\textbf{Algorithm}  				            &           &               & 											& \multicolumn{2}{c?{0.25pt}}{Rel. Speedup}  																&  No. of EM  & Coreset Constr. \\
		\#Clusters	& Data Description 	& Name             & $G$       & $N'$          & Relative Error $\eta$ & \multicolumn{1}{l}{Time}     & \multicolumn{1}{r?{0.25pt}}{Distances} &  Iterations & \& Seeding Time  \\
		\midrule
		\textbf{KDD}         & \multirow[t]{4}{2.5cm}{$N = 145{,}751$, $D = 74$. \\ Protein homology dataset, used in KDD Cup 2004 \citep{KDD2004}}                      
																& $k$-means++           & -         & -             &   0.0\% $~\pm~$ 0.04  &    1.0x       &     1.0x       & 13.0                  & 22\%      \\
		$C = 500$   &               & \textit{var-GMM-S}    &  5        & -             & 0.94\%  $~\pm~$ 0.07  &    6.4x       &    23.4x       & 17.0                  &  1\%      \\
								&               & \textit{LWCS}         & -         & $2^{13}$      & 11.73\% $~\pm~$ 0.18  &   27.8x       &    24.2x       & 10.2                  &  3\%      \\
								&               & \textit{vc-GMM}       &  5        & $2^{13}$      & 10.81\% $~\pm~$ 0.08  &   91.5x       &   361.0x       & 17.9                  & 12\%      \\
		\midrule

		\textbf{RNA}         & \multirow[t]{4}{2.5cm}{$N = 488{,}565$, $D = 8$. \\ RNA input sequence pairs \citep{uzilov2006detection}}
																& $k$-means++           & -         & -             &   0.0\% $~\pm~$ 0.22  &    1.0x       &     1.0x       & 18.0                  & 15\%      \\
		$C = 500$   &               & \textit{var-GMM-S}    &  3 + 1    & -             & -1.59\% $~\pm~$ 0.23  &    1.6x       &    19.2x       & 72.3                  & $<$1\%      \\
								&               & \textit{LWCS}         & -         & $2^{15}$      & 11.24\% $~\pm~$ 0.60  &   21.2x       &    19.8x       & 14.2                  & 3\%      \\
								&               & \textit{vc-GMM}       &  3 + 1    & $2^{15}$      &  7.28\% $~\pm~$ 0.28  &   33.3x       &   329.4x       & 47.8                  & 5\%      \\
		\midrule
		
		\textbf{CIFAR-10}    & \multirow[t]{4}{2.5cm}{$N = 50{,}000$, $D = 3{,}072$. \\ $N = 10.000$ for test set. \\ \citet{krizhevsky2014cifar}, with uniform noise in [0, 1] for cont. data (std. preprocessing)}
																& $k$-means++          & -         & -             &  0.0\%  $~\pm~$ 0.02  &    1.0x        &    1.0x       & 10.7                  & 9\%      \\
		$C = 500$   &               & \textit{var-GMM-S}    &  5        & -             & -0.75\% $~\pm~$ 0.02  &    12.6x       &    16.4x      & 16.7                  & 1\%      \\
								&               & \textit{LWCS}         & -         & $2^{12}$      & 10.34\% $~\pm~$ 0.03  &    17.3x       &    17.6x      & 8.0                   & 2\%      \\
								&               & \textit{vc-GMM}       &  5        & $2^{12}$      & 8.98\%  $~\pm~$ 0.03  &    166.4x      &    207.8x     & 13.4                  & 20\%      \\
		\midrule

		\textbf{SUSY}        & \multirow[t]{4}{2.5cm}{$N = 5{,}000{,}000$, $D = 18$. \\ High-energy physics \citep{baldi2014searching}}
																& $k$-means++           & -         & -             &   0.0\% $~\pm~$ 0.01  &   1.0 x       &    1.0 x       & 47.5                  & 7\%      \\
		$C = 2000$  &               & \textit{var-GMM-S}    &  5        & -             &  0.76\% $~\pm~$ 0.01  &   10.1x       &    83.8x       & 63.7                  & $<$1\%      \\
								&               & \textit{LWCS}         & -         & $2^{15}$      &  9.89\% $~\pm~$ 0.02  &  395.3x       &  381.8x       & 19.3                  & 3\%      \\
								&               & \textit{vc-GMM}       &  5        & $2^{15}$      &  9.55\% $~\pm~$ 0.03  & 1300.4x       & 9964.3x       & 62.8                  & 11\%      \\
		\midrule

		\textbf{SONG}        & \multirow[t]{4}{2.5cm}{$N = 515{,}345$, $D = 90$. \\ Compiled from the Million Song Dataset \citep{bertin2011million}}
																& $k$-means++           & -         & -             &   0.0\% $~\pm~$ 0.01  &    1.0x        &    1.0x       & 11.0                  & 19\%      \\
		$C = 4000$  &               & \textit{var-GMM-S}    &  5        & -             &  0.99\% $~\pm~$ 0.02  &   24.9x        &   79.1x       & 26.2                  & 1\%      \\
								&               & \textit{LWCS}         & -         & $2^{16}$      & 10.42\% $~\pm~$ 0.18  &   10.5x        &   10.3x       & 9.1                   & 1\%      \\
								&               & \textit{vc-GMM}       &  5        & $2^{16}$      &  9.23\% $~\pm~$ 0.07  &  166.6x        &  513.9x       & 22.3                  & 11\%      \\
		\bottomrule
\end{tabular}
\setlength{\tabcolsep}{6pt}
}

\end{table*}

To facilitate comparison to state-of-the-art clustering algorithms, we use the standard quantization error as a measure for clustering quality.
Results are given as the relative error
\begin{equation}
\eta = (\mathcal{Q}_{\mathrm{algo.}} - \mathcal{Q}_{\mathrm{kmpp}}) / \mathcal{Q}_{\mathrm{kmpp}},
\label{eq:eta}
\end{equation}
where $\mathcal{Q}$ denotes the quantization error for the considered approaches ($\mathcal{Q}_{\mathrm{algo.}}$) and for $k$-means++ ($\mathcal{Q}_{\mathrm{kmpp}}$).
Likelihood based measures would be more natural for \textit{vc-GMM} but would hinder comparison with the (at large scales) more frequently used $k$-means-like approaches (also see Suppl.\,\ref{app:AddResults} for additional NMI values).
The algorithms we compare to are:
\textit{AFK-MC$^2$} seeding alone, $k$-means++, \mbox{\textit{var-GMM-S}}, and LWCS followed by $k$-means updates (simply referred to as \textit{LWCS}).
All compared clustering algorithms except for $k$-means++ use AFK-MC$^2$ as fast seeding method with $m = 2$ for KDD, CIFAR-10, SUSY and SONG, and $m = 20$ for RNA.
All algorithms were executed until the same convergence criterion was reached:
We declared convergence if the relative change of the clustering objective fell below a threshold of $10^{-4}$ (Suppl.\,\ref{app:Exp} for details).
More details on the implementation are provided in Suppl.\,\ref{app:ImpDetails}.
Fig.\,\ref{fig:mainFigure} shows clustering times and quantization errors after convergence and for different settings of the algorithms.
In order to investigate the trade-off between execution times and clustering quality, we show different values of $G$ for \textit{var-GMM-S}, different values of $N'$ for \textit{LWCS} and \textit{vc-GMM}, and different values for $m$ for \textit{AFK-MC$^2$} alone.
Quantization errors were for all algorithms computed on the standard test set (for CIFAR-10) or on the full data set (all other benchmarks), and are given relative to the quantization error of $k$-means++, which serves as baseline.
Computational demand was measured in terms of the total number of distance evaluations until convergence (for E-steps, seeding and coreset construction) and is given by Fig.\,\ref{fig:mainFigure}, left column.
Distance evaluations are often used for comparisons \citep{BachemEtAl2016a,ForsterLucke2018} because they are implementation independent and align with the theoretically achievable optimum.
In addition and for the purposes discussed above, we here also show the actual elapsed run time of the algorithms until convergence (Fig.\,\ref{fig:mainFigure}, middle column).
Variational algorithms such as \textit{vc-GMM} make use of more diverse updates than $k$-means.
For \textit{vc-GMM} and \textit{var-GMM-S} overheads include updates of search spaces $\mathcal{G}^{(n)}$ and index sets $\mathcal{K}^{(n)}$ and $\mathcal{G}_{c}$, the computation of approximate posteriors $s_{c}^{(n)}$ from distances and of their use for the M-step.
The question how reduced numbers of distance evaluations for \textit{vc-GMM} also translate into reduced execution times can therefore only be answered by benchmarking the actual run times until convergence.
To measure the elapsed time, we implement all compared algorithms, including seeding and LWCS construction, in C++ source code of equal structure.\footnote{Implementation available at \url{https://bitbucket.org/fhirschberger/clustering/}. For further developments see \linebreak \url{https://github.com/variational-sublinear-clustering}.}.
The elapsed time until convergence was measured running sequentially on a dual Xeon E5-2630 v4 system with multiple random seeds running in parallel.
For our measurements, elapsed time includes all computations required for each algorithm, including seeding, coreset construction and parameter optimization.

Fig.\,\ref{fig:mainFigure} shows results for the different settings as averages over $50$~independent runs with new random seeds.
Time measurements refer to the average time of one sequential execution of the algorithm.
Tab.\,\ref{tab:mainTable} shows more details on the average number of EM iterations until convergence and the time fractions spent for coreset construction and seeding modules.
The performance trade-off for \textit{LWCS} and \textit{vc-GMM} is compared for an error increase of $\sim\!10\%$.

By considering Fig.\,\ref{fig:mainFigure} and Tab.\,\ref{tab:mainTable} it can be observed that \textit{vc-GMM} (Alg.\,1) can strongly reduce execution times in terms of distance evaluations as well as in terms of elapsed time.
The overhead for auxiliary operations and M-steps does (as expected) impact the elapsed time measurements compared to the measured number of distance evaluations.
For smaller scale clustering tasks such as RNA the final speedup is therefore not substantial.
However, like for distance evaluations, elapsed time speedups are very significant especially for large-scale clustering tasks.
For CIFAR-10 with $C=500$ and SONG with $C=4000$, for instance, we observe up to one order of magnitude faster execution times than for (already highly efficient) LWCS-based clustering (with at the same time lower increases in clustering error).

\begin{table}[h]
    \centering
    \caption{
			Quantization error and measured run times (minutes:seconds) of AFK-MC$^2$seeding and variational EM iterations for increasingly large numbers of clusters~$C$ on the 80~Mio.\ Tiny Images dataset.
			Results for a single run.
		}
		\vspace{3pt}
		\label{tab:largescale}
		{
\fontsize{7.5}{9} \sffamily \sansmath
\renewcommand{\arraystretch}{1.1}
\setlength{\tabcolsep}{4pt}
\newcolumntype{?}[1]{!{\vrule width #1}} 

\begin{tabular}[b]{l?{0.1pt}r}
	\toprule
	\multicolumn{2}{c}{\textbf{80 Mio.\,Tiny Images}} \\
	\multicolumn{2}{c}{\citep{TorralbaEtAl2008}} \\
	\midrule
	$N_\mathrm{train}$ 	& $78{,}302{,}017$ 	\\
	$N_\mathrm{test}$ 	& $1{,}000{,}000$ 	\\
	$D$									& $3{,}072$					\\
	\bottomrule
\end{tabular}
\hspace{20pt} 
\begin{tabular}[b]{r?{0.25pt}ccc}
	\toprule
	\multicolumn{4}{c}{\textbf{vc-GMM} ($G = 5+1,\, N' = 10$M) } \\
	\multicolumn{1}{c?{0.25pt}}{$C$} & Q-Error		& Seeding	& EM	\\
	\midrule
	$8{,}192$		& 5.45e+12	& 02:44		& 26:32 \\
	$16{,}384$	& 5.31e+12	& 11:05		& 46:26 \\
	$32{,}768$	& 5.22e+12	& 27:03		& 55:13 \\
	\bottomrule
\end{tabular}
\setlength{\tabcolsep}{6pt}
}

\end{table}

To demonstrate the practical implications, we finally apply \textit{vc-GMM} to a very large scale clustering task using the entire 80~Million Tiny Images dataset \citep{TorralbaEtAl2008}.
We here keep the coreset size to be fixed at $N' = 10$M and measure clustering times (and times of the individual modules) when executing \textit{vc-GMM} on 20~Xeon CPU cores on a single
compute node. As can be observed when considering Tab.\,\ref{tab:largescale}, clustering times for the dataset are in the range of minutes even for large numbers of clusters.
Also observe that by doubling the number of clusters, clustering times do not double. This observation shows a sublinearity of variational EM for clustering \citep[compare][]{ForsterLucke2018}
for the first time in real time and on real data. Finally, also observe that seeding (scaling with $\mathcal{O}(C^2)$) becomes relative to EM more costly, which suggests future work on
the seeding module of \textit{vc-GMM}.

\section{Discussion}

How can we obtain as good as possible clustering results in as short times as possible?
Here we addressed the question with a focus on large and very-large scale clustering problems.
Our main tools were efficient coreset construction methods, efficient seeding, and efficient variational EM, which we combined on the basis of a highly optimized practical algorithm (Alg.\,\ref{alg:vc-GMM}, Sec.\,\ref{sec:AlgRealization}).
In theory and in practice, variational EM and coresets were in this context observed to be ideal counterparts.
Combined they drastically reduce computational demand as well as memory requirements (cf.\ Suppl.\,\ref{app:VarLoop} for an illustration).
While computational demand is the crucial feature for speedups, reduced memory is also important as variational EM would otherwise exceed available hardware memories at large scales.
The complete algorithm, \textit{vc-GMM}, can consequently be applied at scales at which previous approaches, such as \textit{var-GMM-S} \citep{ForsterLucke2018}, have both prohibitive execution times and prohibitive memory demands.
Maybe more importantly, however, \textit{vc-GMM} shows that the substantial theoretical complexity reductions suggested by required numbers of distance evaluations of objective~(\ref{eq:FreeEnergyExplicit}) (but also when no coresets are used \citep{ForsterLucke2018}) indeed translate into substantial practical speedups.
On standard benchmarks (Fig.\,\ref{fig:mainFigure}) we do not only improve on the state-of-the-art by some percents but observe up to an order of magnitude faster clustering times compared to the best recently reported values.
Furthermore, we report with our applications to 80 Mio.\,Tiny Images results on clustering tasks that have previously been considered too computationally demanding for direct clustering.
Indeed, we are not aware of any previous direct fit of a $k$-means-like or GMM model to this very prominent database.

One reason to choose 80~Mio.\ Tiny Images in the first place is that the original contribution \citep{TorralbaEtAl2008} specifically considered the dataset to require {\em non-parametric} approaches for analysis.
Here we now demonstrate that a {\em parametric} density model can be fitted, and that such fits using the here described algorithm are very feasible (see Tab.\,\ref{tab:largescale}).
We do notably not claim that the used Euclidean metric is well suited for image data sets.
The used distance is simply the baseline, and it is used as such in the original paper \citep{TorralbaEtAl2008}, as well as in more recent approaches \citep{OttoEtAl2018}.
As our goal was efficiency increases, Euclidean distance and quantization error are, furthermore, an imposed first choice as means for comparison with competing approaches.
Future work will, however, investigate other metrics using more elaborate GMMs or other mixtures.
Also practical aspects such as parallelization and streaming settings are of future interest.

{
	\setlength{\bibsep}{4.0pt plus 1pt}
	\bibliography{bibliography}
}

\clearpage

\appendix
\setcounter{secnumdepth}{1}
\renewcommand\thefigure{S.\arabic{figure}}
\renewcommand*{\theHfigure}{\thefigure}
\renewcommand\thetable{S.\arabic{table}}
\renewcommand*{\theHtable}{\thetable}
\setcounter{figure}{0}
\setcounter{table}{0}

\section{Derivation of the variational bound for coresets and truncated distributions}
\label{app:FreeEnergy}

The standard approach to finding parameters~$\Theta$ which fit a set of $N$~data points well is to maximize the data log-likelihood $L(\Theta)=\sum_{n=1}^{N} \log\!\big(p(\vec{y}^{\,(n)}|\Theta)\!\big)$.
By inserting the GMM~(\ref{eq:GMMIso}), the maximum likelihood solution~$\Theta^*$ for the clustering of $N$~data points can formally be stated by:
\begin{equation}
\begin{aligned}
& \Theta^* = \underset{\Theta}{\mathrm{argmax}} \big\{L(\Theta)\big\}, \mathrm{with}\\  
&L(\Theta) = \!\sum_{n=1}^{N} \log\!\Bigg(\!\sum_{c=1}^{C} \exp\!\bigg(\!-\!\frac{\|{}\vec{y}^{\,(n)}-\vec{\mu}_c\|{}^2 }{2\sigma^2}\bigg)\!\Bigg) \!+\! B(\sigma),\!
\label{eq:Likelihood}
\end{aligned}
\end{equation}
where $B(\sigma)=-N\log(C)-\frac{ND}{2}\log(2\pi\sigma^2)$.
This likelihood objective requires the evaluation of $NC$ distances; the total cost for computing the objective is therefore $\mathcal{O}(NCD)$.

To derive from this likelihood objective the required variational bounds, we first have to generalize the standard derivation in two ways:
(A)~in order to include the coreset weights, and
(B)~to allow for variational distributions $q^{(n)}(c;\Lambda)$ with `hard' zeros (which will be important for efficiency).

Most of the derivations follow along similar lines as proofs in \citet{Lucke2018}.
The main observation is that the coreset weights do not interfere with the main analytical steps, such that results carry over when coresets are used.

We first derive a variational lower bound of the coreset likelihood (\ref{eq:LikelihoodCore}) along the lines of standard variational EM \citep{JordanEtAl1999}:
\begin{align}
L^{\textrm{core}}(\Theta) & = \sum_{n=1}^{N'} \gamma^{(n)}\log\!\left( p(\vec{y}^{\,(n)}|\Theta) \right) \nonumber\\
&= \sum_{n=1}^{N'} \gamma^{(n)}\log\!\left( \sum_{c=1}^C p(c, \vec{y}^{\,(n)}|\Theta) \right) \nonumber\\
& = \sum_{n=1}^{N'} \gamma^{(n)}\log\!\left( \sum_{c=1}^C q^{(n)}(c;\Lambda) \frac{p(c,\vec{y}^{\,(n)}|\Theta)}{q^{(n)}(c;\Lambda)} \right),
\label{eq:LikelihoodVarCore}
\end{align}

In the last step of (\ref{eq:LikelihoodVarCore}), we introduced strictly positive variational distributions $q^{(n)}(c;\Lambda)$.
To derive a variational bound for truncated distributions $ q^{(n)}(c;\Lambda)=q^{(n)}(c;\mathcal{K},\hat{\Theta})$ with hard zeros for all $c\not\in\mathcal{K}^{(n)}$, we consider the following reformulation:
\begin{align}
q^{(n)}(c;\mathcal{K},\hat{\Theta}) &= \frac{p(c,\vec{y}^{\,(n)}|\hat{\Theta})}{\sum_{\tilde{c} \in \mathcal{K}^{(n)}} p(\tilde{c},\vec{y}^{\,(n)}|\hat{\Theta})} \delta(c\in\mathcal{K}^{(n)}) \nonumber\\
&= \lim_{\epsilon_{n}^{-} \to 0} \tilde{q}^{(n)}(c;\mathcal{K},\hat{\Theta}),
\end{align}
with \vspace{-8pt}
\begin{align}
& \tilde{q}^{(n)}(c;\mathcal{K},\hat{\Theta}) \hspace{-10pt} &&=
\begin{cases}
    q^{(n)}(c;\mathcal{K},\hat{\Theta}) - {\epsilon_{n}^{-}} &\forall c     \in \mathcal{K}^{(n)}\nonumber\\
    q^{(n)}(c;\mathcal{K},\hat{\Theta}) + {\epsilon_{n}^{+}} &\forall c \not\in \mathcal{K}^{(n)}
\end{cases}\\
&&&=
\begin{cases}
    q^{(n)}(c;\mathcal{K},\hat{\Theta}) - {\epsilon_{n}^{-}} &\forall c \in \mathcal{K}^{(n)}\\
    \frac{C'}{C-C'} {\epsilon_{n}^{-}} &\forall c \not\in \mathcal{K}^{(n)}
\end{cases}\!,
\label{eq:qnt}
\end{align}
\vspace{-6pt}
\begin{flalign}
\text{using} && 0 < \epsilon_{n}^{-} < \min\limits_{\,\,c \in \mathcal{K}^{(n)}\!\!} q^{(n)}(c;\mathcal{K},\hat{\Theta}) &&
\end{flalign}
\vspace{-10pt}
\begin{flalign}
\text{and} && \epsilon_{n}^{+} := \frac{|\mathcal{K}^{(n)}|}{C-|\mathcal{K}^{(n)}|} \epsilon_{n}^{-} = \frac{C'}{C-C'} \epsilon_{n}^{-}.&&
\end{flalign}
This reformulation leads to strictly positive variational distributions $\tilde{q}^{(n)}(c;\mathcal{K},\hat{\Theta})$ with $\sum_c \tilde{q}^{(n)}(c;\mathcal{K},\hat{\Theta}) = 1$, that in the limit of ${\epsilon_{n}^{-}}\to 0 $ recover the aimed at truncated distributions $q^{(n)}(c;\mathcal{K},\hat{\Theta})$.
We now insert $q^{(n)}(c;\Lambda)=\lim_{\epsilon_n^- \to 0}\tilde{q}^{(n)}(c;\mathcal{K},\hat{\Theta})$ into Eq.\,(\ref{eq:LikelihoodVarCore}) and use Jensen's inequality to gain a lower bound for truncated distributions:
\begin{align}
L^{\textrm{core}}(\Theta) \!& = \! \mbox{\small $\displaystyle \sum_{n=1}^{N'} \lim_{\epsilon_n^- \to 0} \! \gamma^{(n)}\log\!\bigg( \sum_{c=1}^C \tilde{q}^{(n)}(c;\mathcal{K},\hat{\Theta}) \frac{p(c,\vec{y}^{\,(n)}|\Theta)}{\tilde{q}^{(n)}(c;\mathcal{K},\hat{\Theta})} \bigg)$} \nonumber\\
& \geq \! \mbox{\small $\displaystyle  \sum_{n=1}^{N'} \lim_{\epsilon_n^- \to 0} \gamma^{(n)} \sum_{c=1}^C \tilde{q}^{(n)}(c;\mathcal{K},\hat{\Theta}) \log\!\bigg(\frac{p(c,\vec{y}^{\,(n)}|\Theta)}{\tilde{q}^{(n)}(c;\mathcal{K},\hat{\Theta})} \bigg)$} \nonumber\\
&:= \mathcal{F}(\mathcal{K},\hat{\Theta},\Theta)
\end{align}
Splitting of the sums and evaluation of the limits ${\epsilon_{n}^{-}}\to 0$ then leads to:
\begingroup
\allowdisplaybreaks
\begin{align}
&L^{\textrm{core}}(\Theta) \geq \mathcal{F}(\mathcal{K},\hat{\Theta},\Theta) \nonumber \\
&= \sum_{n=1}^{N'} \gamma^{(n)}  \Bigg[ \! \lim_{\epsilon_n^- \to 0} \!\! \sum_{\,\,c \in \mathcal{K}^{(n)}\!\!} \!\!\! \tilde{q}^{(n)}(c;\mathcal{K},\hat{\Theta}) \log\!\Big(p(c,\vec{y}^{\,(n)}|\Theta)\Big)\nonumber\\[-3pt]
& \phantom{= \sum_{n=1}^{N'} \gamma^{(n)}}\!\! + \lim_{\epsilon_n^- \to 0} \!\! \sum_{c \not\in \mathcal{K}^{(n)}} \!\!\! \tilde{q}^{(n)}(c;\mathcal{K},\hat{\Theta}) \log\!\Big(p(c,\vec{y}^{\,(n)}|\Theta)\Big)\nonumber\\[-3pt]
& \phantom{= \sum_{n=1}^{N'} \gamma^{(n)}}\!\! - \! \!\lim_{\epsilon_n^- \to 0} \!\!\sum_{\,\,c \in \mathcal{K}^{(n)}\!\!}\!\!\! \tilde{q}^{(n)}(c;\mathcal{K},\hat{\Theta}) \log\!\Big(\tilde{q}^{(n)}(c;\mathcal{K},\hat{\Theta}) \Big) \nonumber\\[-3pt]
& \phantom{= \sum_{n=1}^{N'} \gamma^{(n)}}\!\! - \!\lim_{\epsilon_n^- \to 0} \!\!\! \sum_{c \not\in \mathcal{K}^{(n)}}\!\!\! \tilde{q}^{(n)}(c;\mathcal{K},\hat{\Theta}) \log\!\Big(\tilde{q}^{(n)}(c;\mathcal{K},\hat{\Theta}) \Big) \!\Bigg]   \nonumber \\
& = \sum_{n=1}^{N'} \gamma^{(n)} \Bigg[ \! \sum_{\,\,c \in \mathcal{K}^{(n)}\!\!} \! q^{(n)}(c;\mathcal{K},\hat{\Theta}) \log\!\Big(p(c,\vec{y}^{\,(n)}|\Theta)\Big) + 0  \nonumber\\[-3pt]
&\phantom{= \sum_{n=1}^{N'} \gamma^{(n)}}\!\! - \!\sum_{\,\,c \in \mathcal{K}^{(n)}\!\!} \!\!\! q^{(n)}(c;\mathcal{K},\hat{\Theta}) \log\!\Big(q^{(n)}(c;\mathcal{K},\hat{\Theta})\Big) \nonumber\\[-3pt]
&\phantom{= \sum_{n=1}^{N'} \gamma^{(n)}}\!\! - \!\lim_{\epsilon_n^- \to 0} \!\!\sum_{c \not\in \mathcal{K}^{(n)}} \!\! \frac{C'}{C-C'} {\epsilon_{n}^{-}} \log\!\Big(\frac{C'}{C-C'} {\epsilon_{n}^{-}}\Big)\Bigg].
\end{align}%
\endgroup
The second term, which only considers $c\not\in \mathcal{K}^{(n)}$, directly evaluates to zero by definition of the $\tilde{q}^{(n)}(c;\mathcal{K},\hat{\Theta})$.
With \mbox{$\lim_{\epsilon\to 0} \epsilon\log(\epsilon) = 0$}, the last term also disappears and we arrive at the truncated variational bound:
\begin{align}
\mathcal{F}(\mathcal{K},\hat{\Theta},\Theta) &=
 \sum\limits_{n=1}^{N'} \!\gamma^{(n)} \!\!\!\! \sum_{\,\,c \in \mathcal{K}^{(n)}\!\!} \!\!\!\! q^{(n)}(c;\mathcal{K},\hat{\Theta}) \log\!\Big( p(c,\vec{y}^{\,(n)}|\Theta)\Big) \nonumber\\[-3pt]
& + \sum\limits_{n=1}^{N'} \gamma^{(n)} H\big(q^{(n)}(c;\mathcal{K},\hat{\Theta})\big),\label{eq:AppFree}
\end{align}
with
\begin{equation*}
H\big(q^{(n)}(c;\mathcal{K},\hat{\Theta})\big) = - \!\!\!\!\sum_{\,\,c \in \mathcal{K}^{(n)}\!\!} \!\!\!\!q^{(n)}(c;\mathcal{K},\hat{\Theta}) \log\!\Big(q^{(n)}(c;\mathcal{K},\hat{\Theta})\Big).
\end{equation*}
Or, in a more compact form:
\begin{equation}
\mathcal{F}(\mathcal{K},\hat{\Theta},\Theta)
 = \! \sum\limits_{n=1}^{N'} \!\gamma^{(n)} \!\!\!\! \sum_{c \in \mathcal{K}^{(n)}} \!\!\!\! s_{c}^{(n)} \log\!\Big( \frac{p(c,\vec{y}^{\,(n)}|\Theta)}{s_{c}^{(n)}}\Big),
\end{equation}
\begin{flalign*}
\textrm{with }\quad& s_{c}^{(n)} := q^{(n)}(c;\mathcal{K},\hat{\Theta}).&&
\end{flalign*}

This derivation generalizes the derivation for standard likelihoods \citep{Lucke2018} to coreset likelihoods (\ref{eq:LikelihoodCore}).

Based on the lower bound (\ref{eq:AppFree}), the M-step equations~(\ref{eq:GMMMStep}) are derived following the standard procedure:
We first take derivatives of $\mathcal{F}(\mathcal{K},\hat{\Theta},\Theta)$ in (\ref{eq:AppFree}) w.r.t.\ $\vec{\mu}_c$ and $\sigma^2$.
For the derivatives, $\hat{\Theta}$ can be held fixed (which implies that the entropy term can be neglected).
If we demand the derivatives to be zero, we obtain the M-steps (\ref{eq:GMMMStep}).

\section{Derivation of variational E-step}
\label{app:VarEStep}

To derive E-steps for the merged objective~(\ref{eq:FreeEnergyExplicit}), we make use of a number of theoretical results for truncated variational distributions, more specifically:
\begin{enumerate}[label=(\Alph*),topsep=0pt,itemsep=-2pt,partopsep=1ex,parsep=1ex]
	\item We use that $\hat{\Theta}=\Theta$ maximizes $\mathcal{F}(\mathcal{K},\hat{\Theta},\Theta)$ w.r.t.\ $\hat{\Theta}$ while $\mathcal{K}$ and $\Theta$ are held fixed. \vspace{-2pt}
	\item We make use of a simplified functional form of $\mathcal{F}(\mathcal{K},\hat{\Theta},\Theta)$ for $\hat{\Theta}=\Theta$.
	\item We define a partial E-step that increases instead of maximizes the variational bound (as discussed in the main text).
\end{enumerate}

To (A): Considering (\ref{eq:FreeEnergyExplicit}) we can generalize the proof by \citet{Lucke2018} in order to show that $\mathcal{F}(\mathcal{K},\Theta,\Theta)$ is an optimum of the lower bound $\mathcal{F}(\mathcal{K},\hat{\Theta},\Theta)$ if $\mathcal{K}$ and $\Theta$ are held fixed (the coreset weights $\gamma^{(n)}$ do not interfere with the main analytical steps.
The optimization problem that remains for the E-step is consequently the optimization of $\mathcal{F}(\mathcal{K},\Theta,\Theta)=:\mathcal{F}(\mathcal{K},\Theta)$ w.r.t.\ sets $\mathcal{K}$.

To~(B): We can generalize a result for variational bounds with truncated distributions to the coreset-weighted variational bounds (\ref{eq:FreeEnergyExplicit}).
Concretely, we obtain that the functional form of $\mathcal{F}(\mathcal{K},\Theta)$ can be simplified to:
\vspace{-12pt}
\begin{align}
\mathcal{F}(\mathcal{K},\Theta) &=
    \sum\limits_{n=1}^{N'} \gamma^{(n)}
       \log \!\bigg(\!\sum_{\,\,c \in \mathcal{K}^{(n)}\!\!} p(c,\vec{y}^{\,(n)}|\Theta)\bigg).
\label{eq:FreeEnergySimplified}
\end{align}
The functional form of (\ref{eq:FreeEnergySimplified}) is central in solving the optimization problem efficiently.
In contrast to (\ref{eq:AppFree}), the bound (\ref{eq:FreeEnergySimplified}) shows no dependency on an entropy term for the approximate posteriors $s_{c}^{(n)} := q^{(n)}(c;\mathcal{K},\hat{\Theta})$, and indeed no dependency on $s_{c}^{(n)}$ at all.
As $\mathcal{F}(\mathcal{K},\Theta)$ now consists of a single sum with positive coreset weights $\gamma^{(n)}$ and the strictly monotonic logarithm, it is sufficient to find for each $n$ the $C'$ clusters $c$ with the largest joints $p(c,\vec{y}^{\,(n)}|\Theta)$ to maximize (\ref{eq:FreeEnergySimplified}).
The optimization problem on the spaces of size $\binom{\!C}{\,C'}$ for each $n$ is consequently solvable with $\mathcal{O}(N'C)$ computations in total.

We now show, that the simplified variational bound $\mathcal{F}(\mathcal{K},\Theta)$ in Eq.\,(\ref{eq:FreeEnergySimplified}) is indeed a lower bound of the coreset likelihood $L^{\textrm{core}}(\Theta)$ (\ref{eq:LikelihoodCore}) and an upper bound of $\mathcal{F}(\mathcal{K},\hat{\Theta}, \Theta)$ (\ref{eq:FreeEnergyExplicit}), such that
\begin{align}
L^{\textrm{core}}(\Theta) \geq \mathcal{F}(\mathcal{K},\Theta) \geq \mathcal{F}(\mathcal{K},\hat{\Theta}, \Theta).
\label{eq:EnergiesRelations}
\end{align}

Due to the truncated formulation of $\mathcal{F}(\mathcal{K},\Theta)$ and the monotonicity of the logarithm, it is immediately clear, that $\mathcal{F}(\mathcal{K},\Theta)$ is a lower bound of the coreset likelihood:
\begin{align}
L^{\textrm{core}}(\Theta)
&= \sum_{n=1}^{N'} \gamma^{(n)}\log\!\bigg( \sum_{c=1}^C p(c, \vec{y}^{\,(n)}|\Theta) \bigg) \nonumber\\
&\geq  \sum_{n=1}^{N'} \gamma^{(n)}\log\!\bigg( \!\! \sum_{\,\,c \in \mathcal{K}^{(n)}\!\!} p(c, \vec{y}^{\,(n)}|\Theta) \bigg) =: \mathcal{F}(\mathcal{K},\Theta),
\label{eq:LikelihoodRelation}
\end{align}
which becomes tighter, the more probability mass of $p(c, \vec{y}^{\,(n)}|\Theta)$ is covered in the subspace $\mathcal{K}^{(n)}$.

For the relation between $\mathcal{F}(\mathcal{K},\Theta)$ and $\mathcal{F}(\mathcal{K},\hat{\Theta}, \Theta)$, we apply Jensen's inequality to $\mathcal{F}(\mathcal{K},\Theta)$ while using truncated distributions $q^{(n)}(c;\mathcal{K},\hat{\Theta})$ as in (\ref{eq:TruncMain}):
\begin{align}
\mathcal{F}(\mathcal{K},\Theta) &:= \sum_{n=1}^{N'} \gamma^{(n)}\! \log \!\bigg(\!\!\sum_{\,\,c \in \mathcal{K}^{(n)}\!\!} p(c,\vec{y}^{\,(n)}|\Theta)\bigg) \nonumber\\
&= \sum_{n=1}^{N'}\gamma^{(n)}\! \log \!\Bigg(\!\!\sum_{\,\,c \in \mathcal{K}^{(n)}\!\!}\!\! q^{(n)}(c;\mathcal{K},\hat{\Theta}) \frac{p(c,\vec{y}^{\,(n)}|\Theta)}{q^{(n)}(c;\mathcal{K},\hat{\Theta})}\Bigg) \nonumber \\
& \geq \sum_{n=1}^{N'}\gamma^{(n)} \!\!\!\sum_{\,\,c \in \mathcal{K}^{(n)}\!\!}\!\! q^{(n)}(c;\mathcal{K},\hat{\Theta}) \log \!\Bigg( \frac{p(c,\vec{y}^{\,(n)}|\Theta)}{q^{(n)}(c;\mathcal{K},\hat{\Theta})}\Bigg) \nonumber\\
&= \mathcal{F}(\mathcal{K},\hat{\Theta}, \Theta)
\label{eq:FreeEnergyRelation}
\end{align}

The combination of (\ref{eq:LikelihoodRelation}) and (\ref{eq:FreeEnergyRelation}) proves (\ref{eq:EnergiesRelations}) and generalizes the proofs of \citet{Lucke2018} to coreset-weighted data. \hfill $\square$

To show, that the variational bound $\mathcal{F}(\mathcal{K},\hat{\Theta}, \Theta)$~(\ref{eq:FreeEnergyExplicit}) is identical to the simplified form $\mathcal{F}(\mathcal{K},\Theta)$~(\ref{eq:FreeEnergySimplified}) for $\hat{\Theta}=\Theta$, i.e., $\mathcal{F}(\mathcal{K}, \Theta) \equiv \mathcal{F}(\mathcal{K},\Theta, \Theta)$, we proceed as follows:
We insert $s_{c}^{(n)}=q^{(n)}(c;\mathcal{K},\Theta)$ of Eq.\,(\ref{eq:TruncMain}) into the variational lower bound Eq.\,(\ref{eq:FreeEnergyExplicit}), while making sure that $\hat{\Theta}$ is set to $\Theta$.
We then derive:
\begin{align}
\mathcal{F}(\mathcal{K},\Theta, \Theta) & = \! \sum\limits_{n=1}^{N'} \!\gamma^{(n)} \!\!\!\! \sum_{\,\,c \in \mathcal{K}^{(n)}\!\!} \!\!\!\! s_{c}^{(n)} \log\!\bigg( \frac{p(c,\vec{y}^{\,(n)}|\Theta)}{s_{c}^{(n)}}\bigg) \nonumber\\
&=  \sum\limits_{n=1}^{N'} \!\gamma^{(n)} \!\!\! \sum_{\,c \in \mathcal{K}^{(n)}\!} \! \frac{p(c,\vec{y}^{\,(n)}|\Theta)}{\sum_{c' \in \mathcal{K}^{(n)}} p(c',\vec{y}^{\,(n)}|\Theta)} \nonumber\\
&\phantom{=}\times \log\!\bigg(\! \sum_{\,\,c' \in \mathcal{K}^{(n)}\!\!} p(c',\vec{y}^{\,(n)}|\Theta) \bigg) \nonumber\\
& =  \sum\limits_{n=1}^{N'} \!\gamma^{(n)} \log\!\bigg(\!\sum_{\,\,c \in \mathcal{K}^{(n)}\!\!} p(c,\vec{y}^{\,(n)}|\Theta) \bigg) \nonumber\\
&=  \mathcal{F}(\mathcal{K},\Theta)
\label{eq:FreeEnergyEquivalence}
\end{align}
The combination of the equivalence~(\ref{eq:FreeEnergyEquivalence}) for $\hat{\Theta} =\Theta$ and the relation~(\ref{eq:FreeEnergyRelation}) show that $\hat{\Theta} =\Theta$ is a maximum of $\mathcal{F}(\mathcal{K},\hat{\Theta}, \Theta)$ holding $\mathcal{K}$ and $\Theta$ fixed, which is again a generalization of the proofs of \citet{Lucke2018} to coreset-weighted data. \hfill $\square$\\[-2mm]

For further analytical investigations of the properties of these bounds, we now regard the difference between $L^{\textrm{core}}(\Theta)$ and $\mathcal{F}(\mathcal{K},\hat{\Theta}, \Theta)$:
\enlargethispage{\baselineskip}
\vspace{-20pt}
\begin{align}
& L^{\textrm{core}}(\Theta) -  \mathcal{F}(\mathcal{K},\hat{\Theta}, \Theta) \nonumber \\
&= \,\, \sum_{n=1}^{N'} \gamma^{(n)} \log \!\Big( p(\vec{y}^{\,(n)}|\Theta)\Big) \nonumber \\[-3pt]
&\phantom{=} - \sum_{n=1}^{N'}\gamma^{(n)} \!\!\sum_{\,\,c \in \mathcal{K}^{(n)}\!\!}\! q^{(n)}(c;\mathcal{K},\hat{\Theta}) \log \!\Bigg( \frac{p(c,\vec{y}^{\,(n)}|\Theta)}{q^{(n)}(c;\mathcal{K},\hat{\Theta})}\Bigg) \nonumber \\[3pt]
&= \,\, \sum_{n=1}^{N'} \gamma^{(n)} \log \!\Big( p(\vec{y}^{\,(n)}|\Theta)\Big)\nonumber \\[-3pt]
&\phantom{=} - \sum_{n=1}^{N'}\gamma^{(n)} \!\!\sum_{\,\,c \in \mathcal{K}^{(n)}\!\!}\! q^{(n)}(c;\mathcal{K},\hat{\Theta}) \log \!\Big( p(\vec{y}^{\,(n)}|\Theta)\Big) \nonumber \\[-3pt]
&\,\,\, - \sum_{n=1}^{N'}\gamma^{(n)} \!\!\!\!\sum_{\,\,c \in \mathcal{K}^{(n)}\!\!}\!\!\! q^{(n)}(c;\mathcal{K},\hat{\Theta}) \log \!\Big( p(c|\vec{y}^{\,(n)},\Theta)\Big)\nonumber \\[-3pt]
&\phantom{=} + \sum_{n=1}^{N'}\gamma^{(n)} \!\!\!\!\sum_{\,\,c \in \mathcal{K}^{(n)}\!\!}\!\!\! q^{(n)}(c;\mathcal{K},\hat{\Theta}) \log \!\Big( q^{(n)}(c;\mathcal{K},\hat{\Theta})\Big).
\end{align}
Because of the definition of $q^{(n)}(c;\mathcal{K},\hat{\Theta})$, the summation over $c$ of the second term directly evaluates to $\sum_{c \in \mathcal{K}^{(n)}}\! q^{(n)}(c;\mathcal{K},\hat{\Theta}) = 1$ and therefore the first and second term cancel out.
Combination of the third and fourth term then recover coreset-weighted sums of Kullback-Leibler divergences between the distributions $q^{(n)}(c;\mathcal{K},\hat{\Theta})$ and $ p(c|\vec{y}^{\,(n)},\Theta)$:
\begin{align}
&L^{\textrm{core}}(\Theta) -  \mathcal{F}(\mathcal{K},\hat{\Theta}, \Theta) \nonumber \\
&= - \sum_{n=1}^{N'}\gamma^{(n)} \!\!\!\!\sum_{\,\,c \in \mathcal{K}^{(n)}\!\!}\!\!\! q^{(n)}(c;\mathcal{K},\hat{\Theta}) \log\!\Big( p(c|\vec{y}^{\,(n)},\Theta)\Big) \nonumber \\[-3pt]
&\phantom{=}\,\, + \sum_{n=1}^{N'}\gamma^{(n)} \!\!\!\!\sum_{\,\,c \in \mathcal{K}^{(n)}\!\!}\!\!\! q^{(n)}(c;\mathcal{K},\hat{\Theta}) \log \!\Big( q^{(n)}(c;\mathcal{K},\hat{\Theta})\Big) \nonumber \\
&= \sum_{n=1}^{N'}\gamma^{(n)} \!\!\!\!\sum_{\,\,c \in \mathcal{K}^{(n)}\!\!}\!\!\! q^{(n)}(c;\mathcal{K},\hat{\Theta}) \log \!\Bigg( \frac{q^{(n)}(c;\mathcal{K},\hat{\Theta})}{p(c|\vec{y}^{\,(n)},\Theta)}\Bigg) \nonumber \\
&= \sum_{n=1}^{N'}\gamma^{(n)} D_\mathrm{KL}\big(q^{(n)}(c;\mathcal{K},\hat{\Theta}),p(c|\vec{y}^{\,(n)},\Theta)\big) \nonumber \\
&\geq 0,
\label{eq:FreeEnergyDifferences}
\end{align}
where for the identification with the KL-divergence, we again used that \mbox{$\lim_{\epsilon\to 0} \epsilon\log(\epsilon) = 0$}, which allows to expand the sum over all $c$.

The here used M-step for parameter updates (\ref{eq:GMMMStep}) is derived such that $\mathcal{F}(\mathcal{K},\hat{\Theta}, \Theta)$ is increased, and the variational E-step is defined to monotonically increase the lower bound $\mathcal{F}(\mathcal{K}, \Theta)$ (see below).
The used algorithm consequently {\em provably} increases the bound $\mathcal{F}(\mathcal{K}, \Theta)$ in each EM iteration.

\section{Proof of Proposition 1 for coresets}
\label{app:ProofPropOne}

To prove Prop.\,1, we make use of the simplified form of the variational bound for $\hat{\Theta} = \Theta$:
\begin{equation}
	\mathcal{F}(\mathcal{K},\Theta)=\sum_{n} \gamma^{(n)} \log\!\bigg(\!\sum_{\,\,c\in\mathcal{K}^{(n)}\!\!} p(c,\vec{y}^{\,(n)}|\Theta)\bigg).
	\label{EqnFTrunc}
\end{equation}
We now seek for those $\mathcal{K} = \big(\mathcal{K}^{(1)},\dots,\mathcal{K}^{(N')}\big)$ that maximize $\mathcal{F}(\mathcal{K},\Theta)$:
\begin{align}
&&\rlap{$\mathcal{F}(\tilde{\mathcal{K}},\Theta) \geq \mathcal{F}(\mathcal{K},\Theta)$} & \nonumber\\
\Leftrightarrow &&&\sum_{n} \gamma^{(n)} \log\!\bigg(\!\sum_{\,\,c\in\tilde{\mathcal{K}}^{(n)}\!\!} p(c,\vec{y}^{\,(n)}|\Theta)\bigg) \nonumber \\
&&\geq &\sum_{n} \gamma^{(n)} \log\!\bigg(\!\sum_{\,\,c \in \mathcal{K}^{(n)}\!\!} p(c,\vec{y}^{\,(n)}|\Theta)\bigg).
\end{align}
For this maximization, it is sufficient to consider the case where $\forall n$:
\begin{align}
\log\!\bigg(\!\sum_{\,\,c\in\tilde{\mathcal{K}}^{(n)}\!\!} p(c,\vec{y}^{\,(n)}|\Theta)\bigg) \geq \log\!\bigg(\!\sum_{\,\,c \in \mathcal{K}^{(n)}\!\!} p(c,\vec{y}^{\,(n)}|\Theta)\bigg),
\label{eq:FreeEnergyKMaximization}
\end{align}
since for all $n$ where this inequality does not hold, $\tilde{\mathcal{K}}^{(n)}$ can directly be replaced by $\mathcal{K}^{(n)}$ for these $n$-th summands in $\mathcal{F}(\tilde{\mathcal{K}},\Theta)$, which leads to a new increased variational bound, where the inequality~(\ref{eq:FreeEnergyKMaximization}) holds for all $n$.

The optimization w.r.t.\ $\mathcal{K}$ can therefore be regarded as individual optimization of the variational bound w.r.t.\ the $\mathcal{K}^{(n)}$ for each individual corset-weighted data point.
Consequently, this optimization is independent of the individual coreset weights $\gamma^{(n)}$.
Considering the monotonicity of the logarithm, this problem then reduces to finding those $\mathcal{K}^{(n)}$ for each data point $\vec{y}^{\,(n)}$ that have clusters $c\in\mathcal{K}^{(n)}$ with highest cumulative joint probabilities $p(c,\vec{y}^{\,(n)}|\Theta)$.
If we now consider replacement of one $c \in \mathcal{K}^{(n)}$ by a new $c^{\textrm{new}}$ previously not in $\mathcal{K}^{(n)}$, then this increases the variational bound if and only if the joint probability for this replaced cluster $p(c,\vec{y}^{\,(n)}|\Theta) \to p(c^{\textrm{new}},\vec{y}^{\,(n)}|\Theta) $ increases:
\begin{flalign}
 && p(c^{\textrm{new}},\vec{y}|\Theta) &>  p(c,\vec{y}|\Theta) && \nonumber \\
\Leftrightarrow \!\! && \exp\!\Big(-\frac{1}{2\sigma^2}\|\vec{y}-\vec{\mu}_{c^{\textrm{new}}}\|^2\Big) &> \exp\!\Big(-\frac{1}{2\sigma^2}\|\vec{y}-\vec{\mu}_{c}\|^2\Big) && \nonumber \\
\Leftrightarrow \!\! && -\frac{1}{2\sigma^2}\|\vec{y}-\vec{\mu}_{c^{\textrm{new}}}\|^2 &> -\frac{1}{2\sigma^2}\|\vec{y}-\vec{\mu}_{c}\|^2 && \nonumber \\
\Leftrightarrow \!\! && \|\vec{y}-\vec{\mu}_{c^{\textrm{new}}}\| &< \|\vec{y}-\vec{\mu}_{c}\|\,,&&
\end{flalign}
which recovers Prop.\,1. \hfill $\square$

\section{Details of the variational loop of vc-GMM and complexity summary}
\label{app:VarLoop}

\begin{figure*}[tb!]
	\centering
	\resizebox{0.85\textwidth}{!}{
		\begin{adjustbox}{trim=0pt 0pt 0pt 0pt}%
			\input{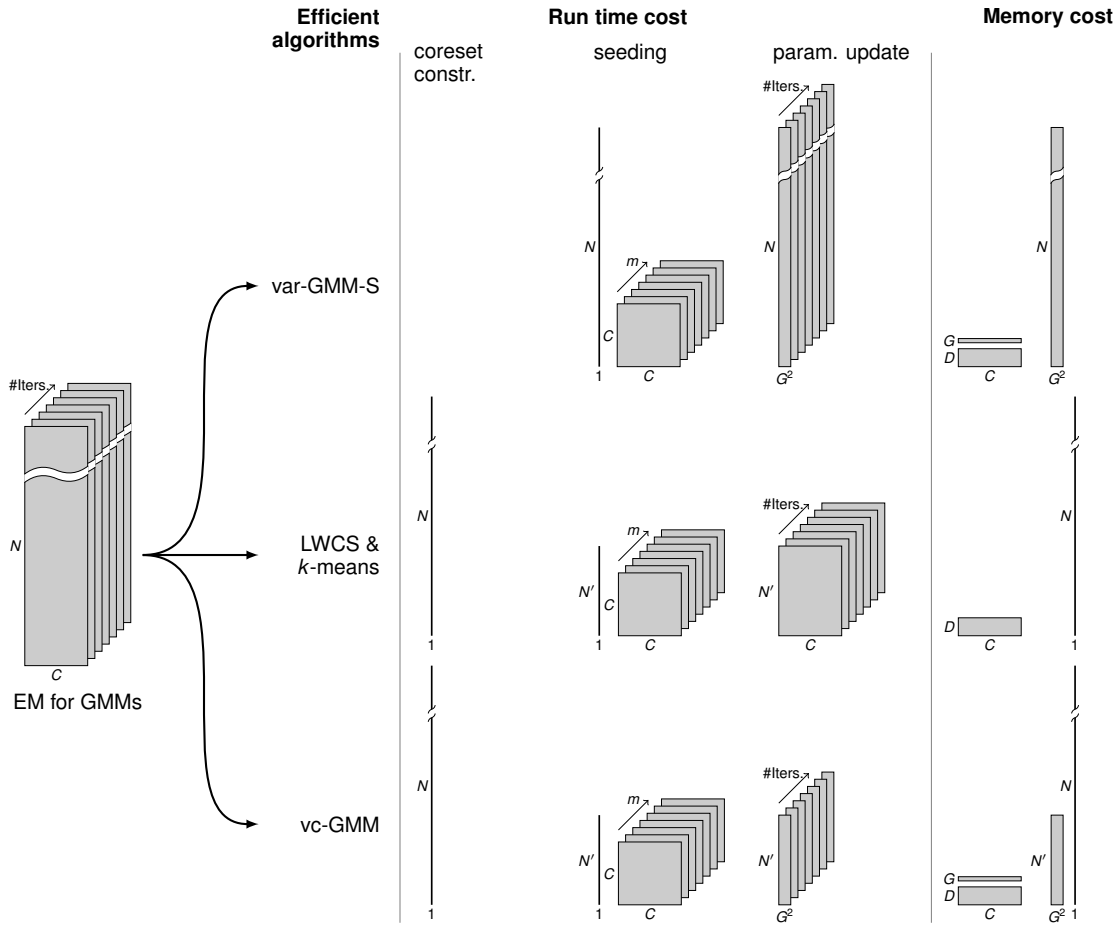}
		\end{adjustbox}
	}
	\caption{
		Graphical illustration of the algorithm's complexities as given in Tab.\ref{tab:complexities}.
		Note, that the figure only shows the qualitative reductions in $N$ and in $C$ respectively.
		In most cases $N$ would be orders of magnitude larger than $C$, and $N'$ would still be a large multiple of $C$.
		Also, in all shown practical use cases of Tab.\,\ref{tab:mainTable} and Tab.\,\ref{tab:largescale}, both the reductions $C\rightarrow G^2$ and $N\rightarrow N'$ were much more drastic than can be visualized properly here.
		The reduction both in run time and memory demand can therefore be much stronger than it appears in the figure.
		The run time cost additionally scales with a factor of $\mathcal{O}(D)$ in all terms, which is for simplicity neglected in the illustration.
		All algorithms are here assumed to use efficient AFK-MC$^2$ seeding, which becomes more dominant the larger $C$ is and the more efficient the algorithm for the parameter updates become.
	}
	\label{fig:complexities}
\end{figure*}

Because of Prop.\,1, the variational loop can be realized for each $n$ of the coreset by using any of the algorithms investigated in \citet{ForsterLucke2018}.
Here we choose the \textit{var-GMM-S} which scales independently of $C$ per iteration and considers more than one `winning' cluster.
As the variational loop only has to be executed for the ${N'}$ data points of the coreset, the run time of the variational E-step of \textit{vc-GMM} has a run-time cost of $\mathcal{O}({N'} G^2D)$ and memory requirement of $\mathcal{O}(CD + N'G^2 + CG + N)$.
Tab.\,\ref{tab:complexities} summarizes the computational complexities of \textit{vc-GMM} as well as the complexities of the approaches we compare to in the numerical experiments.
Fig.\,\ref{fig:complexities} visualizes the same computational costs of the algorithms.
The pseudo-code of the variational loop is given in Alg.\,\ref{alg:vcEStep}.
The nested loops over $c$ and $n$ in the last computational block can (because of the if-condition) be rewritten to scale with $\mathcal{O}({N'} G^2)$ (using $C'=G$)  \citep{ForsterLucke2018}.

\begin{algorithm}[h!]
	\For{$n=1:{N'}$}{    \algro{0.6cm}{\boldmath$\mathcal{O}({N'} C'GD)$}\algBreak
  $\mathcal{G}^{(n)} =\cup_{c\in\mathcal{K}^{(n)}}\mathcal{G}_{c}$;    \algro{0.8cm}{$\mathcal{O}(C'G)$}\algBreak
  \For{$c\in\mathcal{G}^{(n)}$}{                  \algro{0.8cm}{$\mathcal{O}(C'GD)$}\algBreak
      $d^{(n)}_{c} = \|\vec{y}^{\,(n)}\,-\,\vec{\mu}_{c}\|$; \algro{1cm}{$\mathcal{O}(D)$}
  }
  $\mathcal{K}^{(n)}=\{c\,|\,d^{(n)}_{c}$ is among the \algro{0.8cm}{$\mathcal{O}(C'G)$} \\[2pt]
  \hspace{0mm}\hphantom{$\mathcal{K}^{(n)}=\{c\,|$}$C'$ smallest distances$\}$; \\[-1.5mm] \hrulefill
}
\For{$n=1:{N'}$}{ \algro{0.3cm}{\boldmath$\mathcal{O}({N'} C'G)$}\algBreak
  ${c_o^{(n)}} = \argmin_{c\in\mathcal{G}^{(n)}}\big\{d^{(n)}_{c}\big\}$; \algro{0.6cm}{$\mathcal{O}(C'G)$}\\
  $\mathcal{I}_{{c_o^{(n)}}} = \mathcal{I}_{{c_o^{(n)}}} \cup \{n\}$;               \algro{0.6cm}{$\mathcal{O}(1)$} \\ \hrulefill
}
\For{$c=1:C$}{          \algro{0.3cm}{\boldmath$\mathcal{O}({N'} C'G)$}\algBreak
  \For{$n\in\mathcal{I}_c$}{    \algro{0.8cm}{$\mathcal{O}(({N'}/C)C'G)$}\algBreak
    \For{$\tilde{c}\in\mathcal{G}^{(n)}$}{ \algro{1cm}{$\mathcal{O}(C'G)$}\algBreak
        $d_{c\tilde{c}} = d_{c\tilde{c}} + d^{(n)}_{\tilde{c}};$ \algro{1.2cm}{$\mathcal{O}(1)$}\algBreak
        $d_{c\tilde{c}}^\mathrm{\,count} = d_{c\tilde{c}}^\mathrm{\,count} + 1;$             \algro{1.2cm}{$\mathcal{O}(1)$}\algBreak
    }
  } \hrulefill
}
\For{$c=1:C$}{							\algro{0.3cm}{\boldmath$\mathcal{O}({N'} C'G)$}\algBreak
  \For{$n\in\mathcal{I}_c$}{				\algro{0.8cm}{$\mathcal{O}(({N'}/C)C'G)$}\algBreak
    \For{$\tilde{c}\in\mathcal{G}^{(n)}$}{     \algro{1cm}{$\mathcal{O}(C'G)$}\algBreak
        \If{$\mathrm{normalized}_{c\tilde{c}}\neq{}1$}{
          $d_{c\tilde{c}} = d_{c\tilde{c}} / d_{c\tilde{c}}^\mathrm{\,count}$;\algro{1.2cm}{$\mathcal{O}(1)$}\algBreak
          $\mathrm{normalized}_{c\tilde{c}}=1$;\algro{1.2cm}{$\mathcal{O}(1)$}
        }
    }
  }
  $d_{cc}=0;$           \algro{0.8cm}{$\mathcal{O}(1)$}\algBreak
  $\mathcal{G}_{c}= \{\tilde{c}\,|\,d_{c\tilde{c}}$ is among the $G$ \algro{0.8cm}{$\mathcal{O}(({N'} /C)C'G)$} \\[2pt]
  \hspace{0mm}\hphantom{$\mathcal{G}_{c}= \{\tilde{c}\,|$} smallest distances $d_{c:}\}$;  \\[-1.5mm] \hrulefill
}

	\caption{The \textit{vc-GMM} variational loop.\label{alg:vcEStep}}
\end{algorithm}

For the variational loop, we introduce cluster-neighborhoods $\mathcal{G}_{c}$ (with $c \in \mathcal{G}_{c}$) of constant size $G = |\mathcal{G}_{c}|$.
Search spaces to find closer clusters for each data point $\vec{y}^{\,(n)}$ are given by $\mathcal{G}^{(n)} = \cup_{c \in \mathcal{K}^{(n)}}\mathcal{G}_{c}$.
The variational loop then consists of two parts:
First, for each data point $\vec{y}^{\,(n)}$ we compute distances to all clusters $c \in \mathcal{G}^{(n)}$ and select the $C'$ closest clusters to define new $\mathcal{K}^{(n)}$.
Second, we construct sets $\mathcal{I}_{c}$ for each cluster $c = \{1, \dots, C\}$ that hold the indices $n$ of those data points $\vec{y}^{\,(n)}\!$, for which $c$ is the closest found cluster in this iteration.
The sets $\mathcal{I}_{c}$ can therefore be thought of as an {\em estimated} partition of the data set.
If we assume already well converged cluster centers and search spaces that indeed include the closest clusters, then the average data-to-cluster distances of data points in sets $\mathcal{I}_{c}$
{\small
\begin{align}
d_{c c'} \approx \frac{1}{|\mathcal{I}_{c}|} \sum_{n \in \mathcal{I}_{c}} d_{c'}^{(n)}
\end{align}
}%
represent a good estimate for the distances between cluster~$c$ and close-by clusters~$c'$.
In other words:
We estimate the distance $d_{cc'}$  between clusters~$c$ and $c'$ by averaging over distances $d_{c'}^{(n)}$ of clusters~$c'$ to data points $\vec{y}^{\,(n)}$ which lie in close proximity to cluster~$c$.
The distance~$d_{cc}$ between a cluster~$c$ to itself is afterwards manually set to zero.
For more distant clusters~$c'$, where no distance $d_{c'}^{(n)}$ was calculated in this iteration, $d_{c c'}$ is here treated as infinite.
However, since only the distances to the closest clusters are relevant for the update of $\mathcal{G}_c$, a good estimate of close-by cluster distances is sufficient.
And even if the cluster-to-cluster estimates are very coarse, e.g.\ in the beginning of clustering, the definition of the $\mathcal{K}^{(n)}$ updates in Alg.\,\ref{alg:vcEStep} always warrants that the merged objective~(\ref{eq:FreeEnergySimplified}) monotonically increases.
For more details, see \citep{ForsterLucke2018}.

\begin{table*}[bt]
	\centering
  \caption{Computational complexities and memory demands}
	\label{tab:complexities}
	\vspace{3pt}
	{
\fontsize{8}{10} \sffamily \sansmath
\renewcommand{\arraystretch}{1.1}
\setlength{\tabcolsep}{4pt}
\newcolumntype{?}[1]{!{\vrule width #1}}
  \begin{tabular}{l|lll|l}
  \toprule
	& Coreset Constr. &  Seeding  & Param.\ Update & Memory \\
	\midrule
	AFK-MC$^2$                     &  \phantom{iiiii}-   &  $\mathcal{O}(mC^{2}D + ND)$  &  \phantom{iiiii}-      & $\mathcal{O}(CD + N)$\\
	$k$-means with LWCS            &  $\mathcal{O}(ND)$  &  $\mathcal{O}(mC^{2}D + N'D)$  &  $\mathcal{O}(N'CD)$ & $\mathcal{O}(CD + N)$\\
	var-GMM-S                      &  \phantom{iiiii}-   &  $\mathcal{O}(mC^{2}D + ND)$  &  $\mathcal{O}(NG^{2}D)$     & $\mathcal{O}(CD+NG^2+CG)$ \\
	vc-GMM                         &  $\mathcal{O}(ND)$  &  $\mathcal{O}(mC^{2}D + N'D)$  &  $\mathcal{O}(N'G^{2}D)$   & $\mathcal{O}(CD+N'G^2+CG + N)$ \\
	\bottomrule
	\end{tabular}
	\setlength{\tabcolsep}{6pt}
}

\end{table*}

\section{Implementation details}
\label{app:ImpDetails}

Measurements of run times of different algorithms can depend greatly on choices regarding the actual implementation (realization of update rules and used memory structures, programming language, libraries for numerical subroutines, etc).
We therefore chose to use our own C++ implementations of Lloyd's algorithm, $D^{2}$-seeding, AFK-MC$^2$-seeding, and lightweight coreset construction in addition to our C++ implementations of \textit{vc-GMM}.
\textit{var-GMM-S} is trivially realized by \textit{vc-GMM} when no coresets are used (using all data points, and setting all weights $\gamma^{(n)}=1$).
For AFK-MC$^2$ our code follows the published Cython implementation provided by \citet{BachemEtAl2016b}.
For the numerical experiments on 80~Mio.\ Tiny Images we use parallel versions of the algorithms \textit{vc-GMM} and \textit{AFK-MC2} implemented via C++ threads.
To achieve high throughput of vector operations, i.e., distance computations and reductions, we selected the \textit{blaze} library, a high performance C++ math library \citep{blazelib}.
Distance computations are implemented with equally structured code for all algorithms to achieve comparability of run time measurements.
We used double precision in all cases except for the experiments on 80~Mio.\ Tiny Images due to the otherwise significantly larger memory demand.

\section{Details of the numerical experiments}
\label{app:Exp}

For comparisons of the computational cost of different algorithms, we need to define after how many EM iterations parameter updates have converged sufficiently.
In general, we declare convergence of an iterative algorithm when the relative change of the objective it optimizes falls below a predefined threshold.
For \textit{vc-GMM} and \textit{var-GMM-S} these objectives take the form of variational lower bounds (that can be computed efficiently).
Also $k$-means can be interpreted as a variational algorithm \citep{LuckeForster2019} for which a variational lower bound of the likelihood can be defined.
For consistency, we use this bound (which is closely related to the quantization error) for $k$-means.
In the case of $k$-means, the computation of the bound requires additional computations, which we exclude from the measurements of elapsed time.

For all algorithms, we chose a threshold of $\epsilon=10^{-4}$, i.e., the algorithm stops when the relative change of the objective falls below this value.
We observed that a stricter criterion in form of a substantially smaller $\epsilon$, conflicts with the tradeoff between clustering quality and computational cost.
Linear increases in cost with each additional EM iteration would then only yield a marginal improvement in clustering quality.
For example, the difference of the quantization error between a threshold of $10^{-4}$ and $10^{-5}$ for $k$-means++ on the KDD dataset is approximately $0.8$\%.
The number of EM iterations until convergence, however, increases from an average of approx.\ $13$ to $38$.
A smaller threshold of $10^{-3}$ would, on the other hand, result in substantially lower clustering qualities (note the already relatively low number of EM iterations of $k$-means++ for $10^{-4}$).

\begin{figure*}[bt!]
		\begin{adjustbox}{trim=0pt 0pt 0pt 0pt}
			\includegraphics[width=\textwidth]{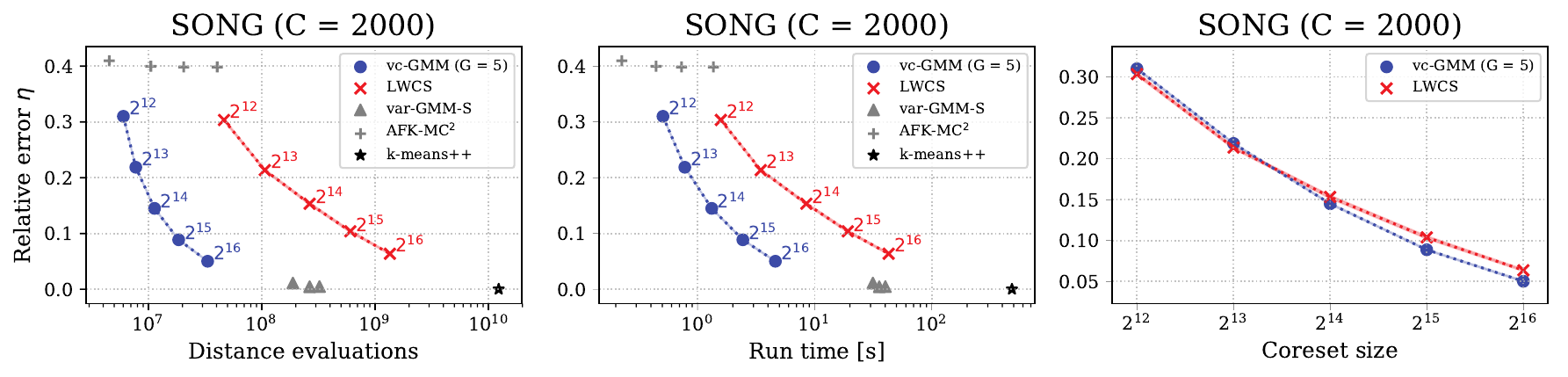}
		\end{adjustbox}
    \caption{Trade-off between relative quantization error of the shown algorithms w.r.t.\ $k$-means++ and speedup (in terms of run time and total number of distance evaluations) on the SONG dataset with $C=2000$ clusters.
		\textit{vc-GMM} and \textit{LWCS} show settings for different coreset sizes (shown in the right column); shaded areas denote the SEM.
    \textit{var-GMM-S} refers to configurations with $G \in \{3+1, 5, 7\}$.
    Seeding with \textit{AFK-MC$^2$} includes chain length of $m \in \{2, 5, 10, 20\}$ (from left to right).}
    \label{fig:Song2k}
\end{figure*}

\begin{figure*}[tb]
		\begin{adjustbox}{trim=0pt 0pt 0pt 0pt}
			\includegraphics[width=\textwidth]{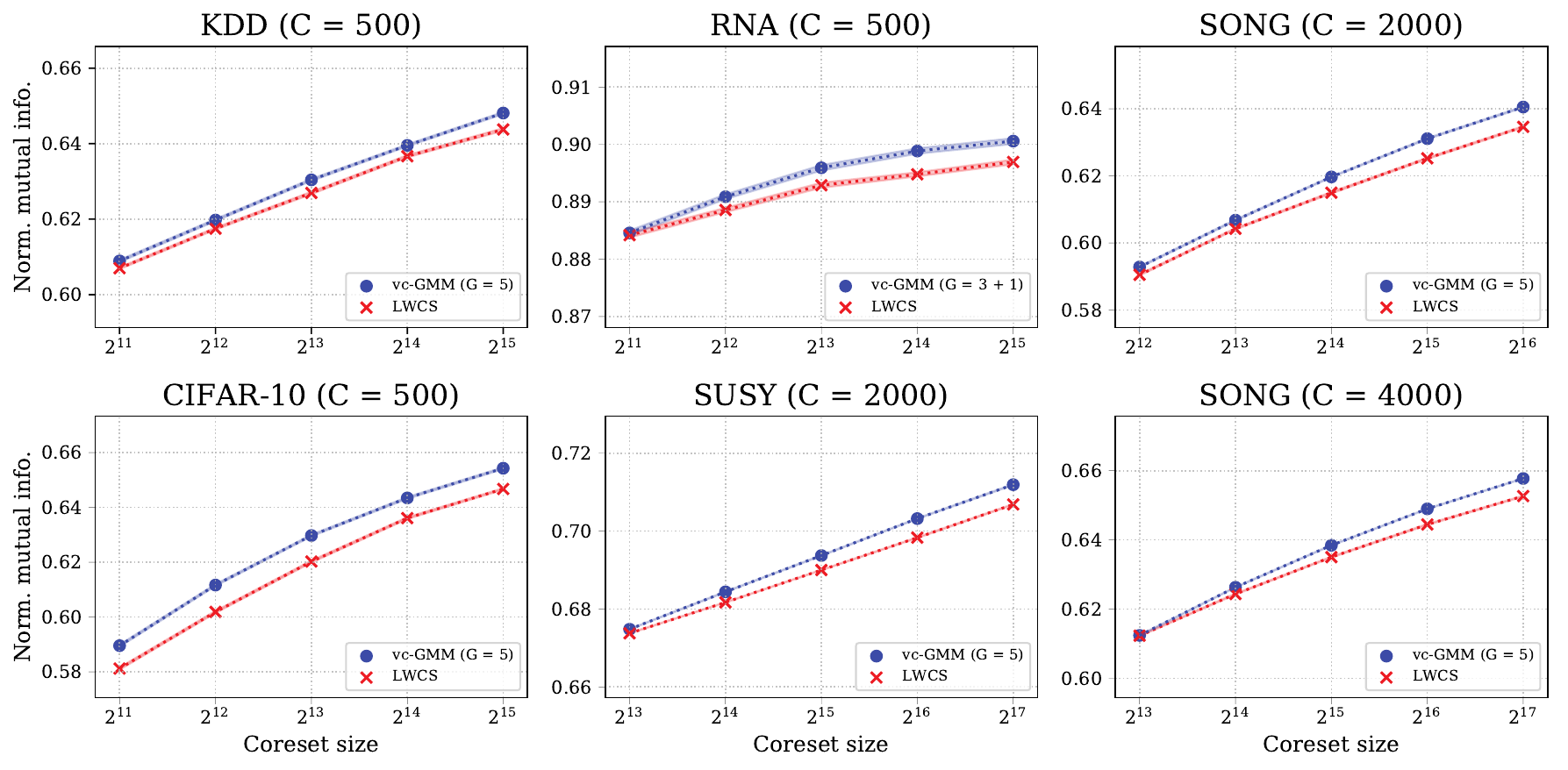}
		\end{adjustbox}
    \caption{NMI score of clustering solutions found by $k$-means on lightweight coresets (\textit{LWCS}) and the merged variational approach \textit{vc-GMM} compared to solutions obtained by $k$-means++ (higher is better). Shaded areas denote the SEM.}
    \label{fig:NMI}
\end{figure*}

Other than the stopping criterion, the number of iterations required for {\em vc-GMM} (and the other used algorithms) until convergence highly depends on the data set and potentially the seeding.
Most iterations were required for variational approaches ({\em vc-GMM} and {\em var-GMM-S}) with small search spaces (small $G$).
Given the strong computational gains per iteration, {\em vc-GMM} with small search spaces was still finally the most efficient approach.

Regarding the choice for $G$ in our experiments, $G=2$ represents the limit of small search spaces, see e.g.\,\citep{ForsterLucke2018}.
For increasingly large $G$ we, in general, observed an increasingly effective increase of the merged objective per iteration but also the computational cost per iteration increases.
Slightly larger than the minimally possible search spaces (i.e., values of $G=3$ to $G=5$) were consistently found to result in the most favorable efficiency vs.\ clustering quality trade-off.
As also observed by \citet{ForsterLucke2018}, the small search spaces of $G=3$ significantly profited from the addition of one random cluster. 
The inclusion of this additional random cluster helps to more quickly improve the initial search space $\mathcal{G}^{(n)}$, as the initial $\mathcal{G}_{c}$ might be unfavorably scattered over far away groups of clusters, with large gaps in between that can not (or only with many iterations) be overcome by regarding immediate cluster neighborhoods alone.

Considering Fig.\,\ref{fig:mainFigure}, the RNA dataset is the data set with the least speedups in elapsed time.
This is mainly due to the low data dimensionality $D$.
A further difference of the RNA dataset compared to the other benchmarks is the in general slower convergence.
\citet{BachemEtAl2016b}, for instance, report that relatively long Markov chains are required for seeding for RNA (while short chain lengths were observed to be sufficient KDD, SONG and SUSY, CIFAR-10).
We here verified this observation (see results for \textit{AFK-MC$^2$}-seeding).
For the RNA and CIFAR-10 datasets we observed that the variational {\em var-GMM-S} algorithm resulted in on average better quantization errors than $k$-means++.
This is presumably the case because of the general tendency of truncated approximations to avoid local optima more effectively \citep{HughesSudderth2016,ForsterEtAl2018}.

For the large scale numerical experiments on the 80~Mio.\ Tiny Images dataset (see Tab.\,\ref{tab:largescale}), we first apply the same preprocessing method as is customary for CIFAR-10, i.e., we add uniform noise in the interval $[0,1]$ to each pixel's color value.
To evaluate quantization errors, we reserve  $N_\mathrm{test} = 1$~Mio.\ data points as a test set and use the remaining data points to construct a lightweight coreset.
We start with $C=2^{13}$ and then double the number of clusters to $2^{14}$ and $2^{15}$. $C=2^{15}$ is a number on the order of the number of non-abstract nouns in the English language \citep{TorralbaEtAl2008}. After convergence of {\em var-GMM-S}, we then computed the quantization error on the full test set without using any approximations, which consequently required $\mathcal{O}(N_\mathrm{test} C D)$ computations.
While computing the quantization error in this way is costly, it has the advantage of being independent of any of the approximations used by the algorithm.
Comparison with future applications of alternative approaches is consequently directly possible.
Coreset construction for 80~Mio.\ Tiny Images takes much longer than the seeding module and EM iterations of vc-GMM, but for the experiments in Tab.\,\ref{tab:largescale}, we have to construct the coreset just once.
The reason for coreset construction being slow is technical: due to the size of the dataset, the implemented coreset algorithm (LWCS) had to stream the data from a hard drive instead of from memory.
Since the execution time for LWCS construction is majorly dominated by this hard drive streaming time, we excluded the coreset construction from the run time measurements in Tab.\,\ref{tab:largescale}.
We run parallel implementations of efficient seeding with \textit{AFK-MC$^2$} and \textit{vc-GMM} using 20 threads on a dual Xeon E5-2630 v4 system with 256\,GB of memory and measure run times for seeding and
variational EM separately. The length of the Markov chains for \textit{AFK-MC$^2$} was set to $m=20$.
All solutions are computed on the same coreset and we only compute a single run for each configuration.

When it was first suggested, {\em var-GMM-S} used initial E-steps before the model parameters were updated for the first time \citep{ForsterLucke2018}.
For the 80~Mio.\ Tiny Images dataset we opted for $5$ initial E-steps without optimizing for this parameter due to the comparably large run times.
For the datasets considered in Fig.\,\ref{fig:mainFigure}, we evaluated the effect of different numbers of initial E-steps, but found that such E-steps did not in general reliably result in any or in significant performance gains.
The exception was again the RNA dataset.
Here, initial E-steps were observed to be favorable.
For \textit{var-GMM-S} we find that for $G = 3 + 1$, initial E-steps improve the average quantization error from $1.73 \cdot 10^{6}$ to $1.65 \cdot 10^{6}$ (four E-steps), and for $G = 5$ from $1.73 \cdot 10^{6}$ to $1.67 \cdot 10^{6}$ (two E-steps).

\section{Additional numerical results}
\label{app:AddResults}

Fig.\,\ref{fig:Song2k} provides further results on the SONG data set for the standard setting of $C=2000$ clusters additional to the higher scale setting of $C=4000$ that we already showed in Fig.\,\ref{fig:mainFigure}.
To better evaluate the trade-off between clustering quality and speedup, Fig.\,\ref{fig:NMI} furthermore shows for all data sets the normalized mutual information (NMI) of the hard partitions of \textit{vc-GMM} (given by the MAP estimate) and \textit{LWCS} with respect to the partitions found by standard $k$-means++.
The evaluation is done on the test data set for CIFAR-10 or on the full data set for all other benchmarks.
For all coreset sizes \textit{vc-GMM} shows to better reproduce the $k$-means++ partitions than the LWCS version of $k$-means.
The respective speedups of \textit{vc-GMM} and \textit{LWCS} are the same as in Fig.\,\ref{fig:mainFigure} (left and middle column), which again shows the superior trade-off between speed and clustering quality (now in terms of NMI) given by \textit{vc-GMM}.

\end{document}